%% file: owloop-th.tex
\documentclass{./template/tlp}
\pdfoutput=1 
\usepackage{float}
\usepackage{amssymb}
\usepackage[utf8]{inputenc}
\usepackage[T1]{fontenc} 
\usepackage{mathrsfs}
\usepackage{multirow}
\usepackage{centernot}
\usepackage{mathtools} 
\usepackage{tikz}
\usepackage{setspace}
\usepackage{xspace}
\usepackage[english]{babel}
\usepackage{cmap}
\input{glyphtounicode}
\usepackage[linesnumbered,algoruled,vlined,algonl]{algorithm2e}
\DontPrintSemicolon
\makeatletter
    \usepackage{etoolbox}
    \newcounter{algoline}
    \newcommand\Numberline{\refstepcounter{algoline}\nlset{\thealgoline}}
    \AtBeginEnvironment{algorithm}{\setcounter{algoline}{0}}
\newcounter{pseudocode}
\newenvironment{pseudocode}[1][htb]
  {
   \let\c@algocf\c@pseudocode
   \begin{algorithm}[#1]%
   \LinesNumberedHidden
  }{\end{algorithm}}
\makeatother
\makeatletter
\newcounter{examplecode}
\newcommand{\ALGOINDENT}{0.5em}
\IncMargin{\ALGOINDENT}
\newenvironment{examplecode}[1][htb]
  {
   \let\c@algocf\c@examplecode
   \begin{algorithm}[#1]%
  }{\end{algorithm}}
\makeatother
\IncMargin{.3em}
\SetAlCapNameFnt{\footnotesize}
\SetAlCapFnt{\footnotesize}
\usepackage[hidelinks]{hyperref}
\hypersetup{hidelinks,%
            pdfauthor={Luca Buoncompagni and Fulvio Mastrogiovanni},%
            pdftitle={OWLOOP: Interfaces for Mapping OWL Axioms into OOP Hierarchies},%
            pdfsubject={An architecture that wraps the OWL language to be compliant with OOP.},%
            pdfkeywords={Ontology Web Language (OWL), Object Oriented Programming (OOP), Application Programming Interface (API) languages, Software Architecture, Unified Modeling Language (UML).},%
            pdfproducer={LaTeX},%
            pdfcreator={pdfLaTeX}
}
\usepackage{xspace}
\input{./numenclature.tex}
\input{./figure/figures_setup.tex}
\setboolean{shouldegeneratefigure}{false}
\usepackage{scalerel}%
\usetikzlibrary{svg.path}%
\definecolor{orcidlogocol}{HTML}{A6CE39}%
\tikzset{%
    orcidlogo/.pic={%
        \fill[orcidlogocol] svg{M256,128c0,70.7-57.3,128-128,128C57.3,256,0,198.7,0,128C0,57.3,57.3,0,128,0C198.7,0,256,57.3,256,128z};%
        \fill[white] svg{M86.3,186.2H70.9V79.1h15.4v48.4V186.2z}%
        svg{M108.9,79.1h41.6c39.6,0,57,28.3,57,53.6c0,27.5-21.5,53.6-56.8,53.6h-41.8V79.1z M124.3,172.4h24.5c34.9,0,42.9-26.5,42.9-39.7c0-21.5-13.7-39.7-43.7-39.7h-23.7V172.4z}%
        svg{M88.7,56.8c0,5.5-4.5,10.1-10.1,10.1c-5.6,0-10.1-4.6-10.1-10.1c0-5.6,4.5-10.1,10.1-10.1C84.2,46.7,88.7,51.3,88.7,56.8z};%
    }%
}%
\newcommand\orcidicon[1]{\href{https://orcid.org/#1}{%
    \ifthenelse{\boolean{shouldegeneratefigure}}{%
        \mbox{\scalerel*{\begin{tikzpicture}[yscale=-1,transform shape]\pic{orcidlogo};\end{tikzpicture}}{|}}%
    }{%
        \centering
        \includegraphics[width=2.5mm]{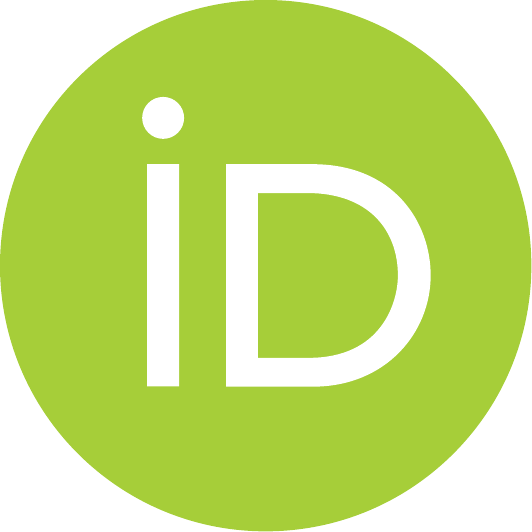}%
    }}%
}%
\begin{document}
\lefttitle{Luca Buoncompagni and Fulvio Mastrogiovanni}
\title[OWLOOP: Interfaces for Mapping OWL Axioms into OOP Hierarchies]
      {OWLOOP: Interfaces for Mapping\\OWL Axioms into OOP Hierarchies%
        {\textsuperscript{$\,\star$}}\let\thefootnote\relax\footnotetext{\hspace{-.8em}\textsuperscript{$\;\star$}%
            This manuscript details the implementation of the OWLOOP API. 
            A simplified (and \emph{citable}) presentation of our API has been published in the SoftwareX Elsevier journal with the title \emph{``OWLOOP: A modular API to describe OWL axioms in OOP objects hierarchies''} (DOI: \url{https://doi.org/10.1016/j.softx.2021.100952}).%
        }%
      }%
\begin{authgrp}
\author{\sn{Luca} \gn{Buoncompagni}\textsuperscript{\orcidicon{0000-0001-8121-1616}} and 
        \sn{Fulvio} \gn{Mastrogiovanni}\textsuperscript{\orcidicon{0000-0001-5913-1898}}}
\affiliation{\vspace{.4em}%
             Department of Informatics, Bioengineering, Robotics and Systems Engineering,\\
             University of Genoa, Via Opera Pia 13, 16145, Genoa, Italy.\\[.4em]
             \href{mailto:buon_luca@yahoo.com}{luca.buoncompagni@edu.unige.it}\\[.4em]
             Manuscript published the 14\textsuperscript{th} April 2024%
            }%
\end{authgrp}
\maketitle
\thispagestyle{empty} 

\begin{abstract} 
The paper tackles the issue of mapping logic \emph{axioms} formalised in the Ontology Web Language (OWL) within the Object-Oriented Programming (OOP) paradigm. 
The issues of mapping OWL axioms hierarchies and OOP objects hierarchies are due to OWL-based reasoning algorithms, which might change an OWL hierarchy at runtime; instead, OOP hierarchies are usually defined as static structures. 
Although programming paradigms based on \emph{reflection} allow changing the OOP hierarchies at runtime and mapping OWL axioms dynamically, there are no currently available mechanisms that do not limit the reasoning algorithms. 
Thus, the factory-based paradigm is typically used since it decouples the OWL and OOP hierarchies. 
However, the \emph{factory} inhibits OOP polymorphism and introduces a paradigm shift with respect to widely accepted OOP paradigms. 
We present the OWLOOP API, which exploits the \emph{factory} to not limit reasoning algorithms, and it provides novel OOP interfaces concerning the axioms in an ontology.
OWLOOP is designed to limit the paradigm shift required for using ontologies while improving, through OOP-like \emph{polymorphism}, the modularity of software architectures that exploit logic reasoning. 
The paper details our OWL to OOP mapping mechanism, and it shows the benefits and limitations of OWLOOP through examples concerning a robot in a smart environment.
\end{abstract}


\section{Introduction}
\label{sec:introduction}

The Ontology Web Language (OWL) is a semantic formalism standardised by the \cite{OWL2Overview}.
It structures knowledge for a domain of interest through symbols about \emph{things}, groups of things, and relations between things. 
Such a knowledge is structured through hierarchies that formally represent the concepts of interest for an application.
OWL-based hierarchies are subjected to consistency checking performed by a logic \emph{reasoner}.
The latter is an algorithm that makes explicit the qualitative and quantitative knowledge implicitly encoded in an ontology.

Ontologies are effective in many applications, which include: 
\emph{static representations}, \eg \cite{abadieowl} formalised the contents of a dataset, and \cite{7968051} the components of an architecture;
\emph{qualitative and quantitative models}, \eg \cite{stevens2007using} and \cite{s2017Artificiala} used ontologies for biological and cognitive systems, respectively; 
\emph{context-based systems}, \eg to recognise human activity in a smart home, \cite{meditskos2016metaq}, and for driving assistance, \cite{6856509};
\emph{semantic web}, \eg \cite{api_ontoAlign} proposes an API for ontology alignment; 
\emph{recommendation systems}, \eg \cite{tarus2018knowledge};
\emph{task planning and scheduling}, \eg \cite{kockemann2018integrating} and \cite{8441117}, respectively;
\emph{support decision makers}, \eg \cite{11928-7_81};
\emph{human-machine interaction}, \eg to ground dialogues, \cite{8243748}, to manage multi-robots systems, \cite{moon2019pddl},
\emph{to drive reinforcement learning}, \cite{48036-7_21};
and \emph{actions explanation}, \eg \cite{02671-4_24}.

It is important to note that such a variety of applications must satisfy very different specifications and that an ontology is meant to be used in close synergy with other software tools, as highlighted by \cite{h2001Ontologybased}; thus, integration is a key aspect.
Systems might involve static knowledge structures that do not require frequent updates and reasoning processes or on-demand computation, \eg in the semantic web paradigm.
On the other hand, there might be scenarios requiring reasoning within real-time specifications, which involve \emph{dynamic} ontologies exploited by intelligent agents to accomplish certain tasks.
Although our contribution addresses both scenarios, we focus more on the latter since we experienced usability issues in related applications, \eg a robot that learns spatial configurations while interacting with a human supervisor, and for recognising contextualised human activities in smart environments, as reported by \cite{oldSIT} and \cite{mon}, respectively.

The issues we experienced are due to the complexity of intelligent agents, which require well-designed software architectures to implement their capabilities, \eg perceiving the environment, being aware of the context, understanding user intentions, taking decisions, and acting accordingly. 
As addressed by \cite{armor_ws}, reasoning on the knowledge of such a complex system requires a \emph{knowledge-centred} architecture, where software components generate and consume the knowledge collected in centralised structures, which are subjected to a reasoning algorithm that must be synchronised among all components.
However, using OWL ontologies within those architectures seems to be so difficult that developers prefer other types of data structures, \eg relational databases, even if they are less powerful as far as semantic reasoning is concerned. 

\cite{s2018Objectoriented} adduced many reasons why developers are reluctant to use ontologies in their applications, including the not homogeneity of software tools and the need for a paradigm shift to exploit OWL ontologies.
Therefore, we argue for better integration of OWL ontology and reasoners with a popular paradigm that would decrease developing effort.
In particular, we aim for an Application Programming Interface (API) concerning a modular representation of knowledge, which is intrinsically extendible and, therefore, flexible. 
Also, we aim for an API that implements an intuitive paradigm for sharing semantic knowledge among the components of an architecture.
Remarkably, software components should be implemented for general purposes, and their design might not be fully under the control of developers, \ie third party libraries.
Therefore, the flexibility of the components using OWL knowledge is a relevant aspect.
 
If the OWL knowledge is interfaced with software components as OOP-based hierarchies of objects able to provide their semantics (e.g., getting the type of an instance or its properties), we would achieve modularity features within the above rationale.
Hence, our objective is to design an OWL to OOP mapping strategy that does not limit the reasoning capabilities and provides developers with an OOP-based interface for exploiting ontologies and reasoners in their software components.

The paper motivates our contribution with respect to related work in the following Section.
Then, Section~\ref{sec:owlPrimer} introduces the OWL formalism, whereas Sections~\ref{sec:OWLOOPDescriptor}--\ref{sec:implementation} describe our novel OWL to OOP mapping strategy.
Section~\ref{sec:examples} provides some examples in a scenario where a robot moves within a smart environment, whose discussion leads to the conclusions of the paper.

\section{Background}
\label{sec:related_work}

OWL ontologies come with several tools and frameworks.
For instance, \cite{m2011OWL} proposed the OWL-API, while \cite{JenaAPI} presented the Jena API, which can be used to develop software that manipulates and queries ontology-based data structures.
\cite{2008SPARQLDL} discussed query engines based on the SPARQL Protocol and the RDF Query Language.
Typically, OWL-based APIs support several reasoners, such as Pellet, proposed by \cite{e2007Pellet}, and Hermit, by \cite{glimm2014hermit}, which can support incremental reasoning, as discussed by \cite{grau2010incremental}.


OWL ontologies encode knowledge through hierarchies containing symbols derived from a list of logic \emph{axioms}, which can either be asserted by software components or inferred by reasoners.
Since all tools and frameworks related to OWL \emph{write} and \emph{read} logic axioms (\ie they assert knowledge and make queries), it is important to interface OWL symbols with a data structure used by software components and mapped as OWL axioms.
We want to design such a data structure through a hierarchy of OOP objects because OOP is a modular paradigm well integrated with software components and provides an intuitive representation of the knowledge in the ontology.
In addition, OOP objects allow defining computational \emph{methods}, which can reflect the semantic relations among axioms in the ontology, \eg getting all instances belonging to a queried concept.

However, even if OWL and OOP hierarchies are similarly structured there are no trivial differences, which have been discussed by \cite{h2006Semantic}.
For instance, one issue is related to the fact that an ontology allows a \emph{class} to derive from multiple classes, and another issue concerns \emph{attributes} that can exist without being referred to a class. 
\cite{s2018Objectoriented} highlighted two paradigms to address the issues of mapping OOP objects and OWL axioms, namely the \emph{active} or \emph{passive} mechanism, also known as ontology-oriented programming and API-based strategy, respectively.

\begin{table}%
    \caption{The list of acronyms concerning OWL and OWLOOP expressions.}%
    \label{tab:acronyms}%
    \input{./table/acronym.tex}\end{table}%

\subsection{Active OWL to OOP Mapping Strategy}
An active strategy requires building an ontology starting from its syntactic form, \eg the eXtensible Markup Language (XML), with the aim of coding statements in the target programming language. 
The resulting ontology is \emph{executable}, and it includes OWL-related features (\eg those concerning reasoning capabilities) as part of the OOP hierarchy implementing the system.

Active mapping has the benefits to provide a native OOP structure for an ontology, but the drawback is that all the issues related to OWL to OOP mapping need to be solved.
Those issues are far from trivial if static OOP hierarchies are used but they become tractable if the \emph{reflection} paradigm is adopted to design dynamic OOP hierarchies, as discussed by \cite{demers1995reflection}.
However, the comparison presented by \cite{s2018Objectoriented} highlights that all the studied mechanisms that implement an active mapping limit reasoning capabilities since they are not able to map all possible OWL semantics within OOP objects.
In particular, \cite{s2018Objectoriented} compared the mechanisms proposed by: \cite{s2006OWLFull}, \cite{stevenson2011sapphire}, \cite{baset2016ontojit}, \cite{j2017Owlready} and \cite{m2017Essence}.

\subsection{Passive OWL to OOP Mapping Strategy}
A passive mapping loads an ontology into memory, where it can be manipulated by external processes, \eg reasoners as well as software components.
In this case, the ontology itself is not executable, but it is reduced to a centralised knowledge structure, \eg an Abstract Syntax Tree (AST), to be interfaced with each component that needs to use the ontology.

This strategy is the most used to deploy ontologies in architectures since it does not limit OWL reasoner capabilities, but it leads to modularity issues and a paradigm shift, as introduced in the previous Section.
In particular, a passive strategy is usually based on the \emph{factory} paradigm, which allows to decouple OWL and OOP hierarchies and avoid issues due to OWL and OOP differences.
However, as highlighted by \cite{4222592}, the factory paradigm leads to development drawbacks since it would not interface OWL knowledge in an OOP fashion; instead, it only encodes data in some predefined OOP objects.
For instance, with the factory paradigm is not possible to extend an object provided by the factory without modifying the factory itself. 
Also, objects provided by the factory are independent of each other even if they are semantically related in the ontology, \ie the objects provided by the factory do not implement any OOP method except the one returning the value of its constant attributes.

\subsection{Contribution}
We present a novel API called OWLOOP, which allows a software component to manipulate axioms and query the reasoner through OOP-based object hierarchies.
OWLOOP implements a passive OWL to OOP mapping for not limiting reasoning capabilities and being compatible with most developed systems based on OWL ontologies. 
Indeed, OWLOOP is based on the factory paradigm since it wraps around OWL-API, but it also provides a novel abstraction layer designed to hide the factory paradigm, and it provides developers with an OOP-like interface to the ontology.
In particular, from the perspective of software components, OWL axioms are encoded as OOP hierarchies defining classes associated with certain OWL semantics (\eg the logic concept of a room), which instances are encoded as OOP attributes (\eg the representation of specific rooms), whereas OOP methods mirror the semantic relationship among instances stored in the ontology (\eg rooms connected through a door).

OWLOOP provides modularity features while avoiding the paradigm shift, without the need to solve all OWL to OOP mapping issues.
However, since OWLOOP does not implement a native OOP interface as the active mapping strategy, it might require the design of specific OOP classes that depend on the knowledge in a particular ontology.
Although OWLLOP defines this dependence in an extendible fashion -- which is not assured with the factory paradigms only -- the system maintainability would be less immediate with respect to an active mapping strategy.
Nevertheless, OWLOOP provides OOP-based \emph{templates} to make trivial the implementation of the objects that depend on the ontology. 
In particular, these objects would inherit from the templates the methods that read OOP-based knowledge and write OWL axioms through the factory.
This paper details such templates to span all OWL semantics and, for showing purposes, we implemented a prototyping API\footnote{\label{nt:github}OWLOOP implementation is available at \url{https://github.com/buoncubi/owloop}.} that concerns the most used OWL semantics.

The paper discusses OWLOOP through examples concerning a scenario where a robot moves in a smart environment.
With those examples, we show the following OWLOOP features. 
\begin{enumerate}
    \item[$I$.]   OWL knowledge in the ontology can be encoded into non-static attributes of an OOP object by \emph{reading} axioms with a given semantic.
    \item[$II$.]  Dynamic knowledge encoded as OOP attributes can be stored into an OWL ontology by \emph{writing} axioms with a specified semantic at runtime.
    \item[$III$.] Through reading and writing, software components can be interfaced with the ontology and perform OWL reasoning in a synchronised manner. 
    \item[$IV$.]  The amount of knowledge mapped through reading and writing is customizable based on the application.
    \item[$V$.]   An OOP object encoding OWL knowledge provides methods to \emph{get} (and \emph{set}) other OOP objects that encode further knowledge based on the semantic relations specified in the ontology.
    \item[$VI$.]  An OOP object encoding OWL knowledge can be extended to inherit functionalities from other OOP objects based on \emph{polymorphism}.
    \item[$VII$.] OWLOOP always allows accessing the factory-based interface to the ontology, \ie OWLOOP is compatible with all software based on the OWL-API.
\end{enumerate}

\begin{figure}[t!]
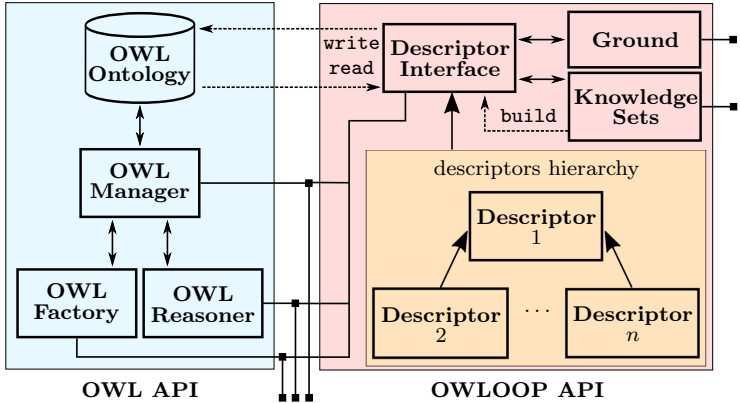
%
    \FIGabstract%
    \caption{An overview of OWLOOP, where single-ended arrows are OOP extensions, squared arrows indicate the API endpoints, double-ended arrows are dependencies, and dashed arrows are OOP-based methods. 
             OWLOOP extends the OWL-API to implement a hierarchy of descriptors to read and write ontological knowledge and build new descriptors.
    }%
    \label{fig:abstract}%
\end{figure}%
                
\section{Overview}
\label{sec:overview}
Under a simplified perspective, an OWL ontology is a structure containing a set of axioms, i.e., statements that specify what holds true in a domain of interest.
Using the OWL-API, axioms can be asserted in an ontology, retrieved from ontology and inferred through reasoning. 
Figure~\ref{fig:abstract} shows, on the left-hand side, a depiction of its architecture composed of an OWL \emph{ontology} module interfaced with an OWL \emph{manager}, which relies on an OWL \emph{factory} and an OWL \emph{reasoner}.
The OWL-API has three endpoints: one is directly related to the OWL manager (\eg for loading and saving an ontology), whereas the others are related to the OWL factory and to the OWL reasoning modules for \emph{asserting} and \emph{inferring} axioms, respectively.
It is worth noting that the knowledge provided by the OWL-API are copies of axioms that are independent of the ontology, and they are not related to each other. 

OWLOOP extends the OWL-API by using its interface to provide a hierarchy of OOP objects, which is customisable for each application (right-hand side of Figure~\ref{fig:abstract}).
Each object in the hierarchy must be a descriptor, \ie it needs to realise the \emph{descriptor interface}, which describes OWL axioms with a specified semantic.
A descriptor collects some axioms in a \emph{knowledge set} while inheriting the \emph{read} and \emph{write} methods, which can be used to synchronise the state of the knowledge set with the ontology.
Also, each descriptor inherits the \CODE{build} methods, which allow instantiating a new descriptor with a knowledge set populated based on semantic relationships in the ontology.

\begin{figure}
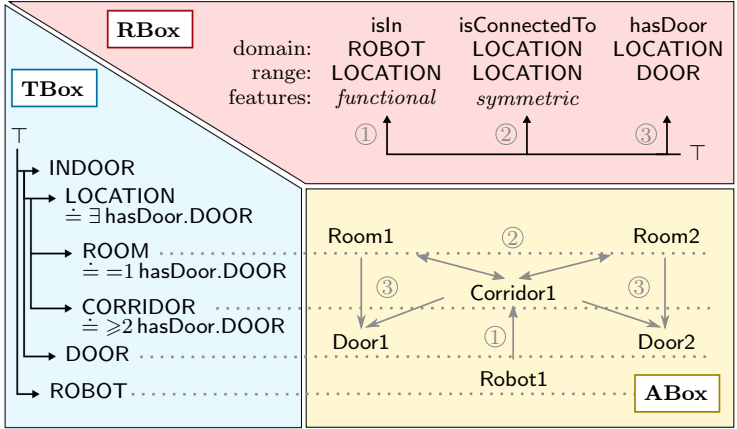

    \FIGontoExample
	\caption{The axioms of a simple ontology used as an example. 
	         It shows the hierarchy of classes and properties in the TBox and RBox, respectively. 
	         It also shows individuals in the ABox and the related assertions in grey, \ie their classification (point-line) and properties (arrows).
	}%
	\label{fig:owloop:example}%
\end{figure}

\begin{table}%
    \caption{The glossary of the symbols used in this paper.}%
    \label{tab:glossary}
    \input{./table/glossary.tex}\end{table}%

A descriptor represents OWL axioms related to a given OWL entity in the ontology called \emph{ground}.
All the methods of a descriptor are applied with respect to the ground, while their results are stored in the knowledge set, and both are provided to OWLOOP users.
OWLOOP provides \emph{primitive} descriptors for different semantics, which can be easily aggregated to design \emph{compound} descriptors having a ground and multiple knowledge sets, each with a specified type of axioms.

The next Section presents the OWL formalism, while the rest of the paper discusses the OWL to OOP mapping implemented in OWLOOP and its usage.
In particular, Section~\ref{sec:OWLOOPDescriptor} introduces descriptor-related interfaces.
Section~\ref{sec:mapping} discusses the details of our OWL to OOP mapping, whereas Section~\ref{sec:implementation} focuses on primitive descriptors and their combinations.
Throughout the paper, we support our discussion with an example involving a robot deployed in a smart environment, as sketched in Figure~\ref{fig:owloop:example}.
That example concerns a topological map that dynamically describes the states of the robot and sensors.

\section{An OWL Primer}
\label{sec:owlPrimer}
OWL has three increasingly-expressive sublanguages, namely OWL-Lite, OWL-DL, and OWL-Full. 
Among them, the Description Logic (DL) formalism lies within the decidable fragment of the family of languages based on First-Order Logic, as described by \cite{f2017Introduction}, and it is one of the most used OWL sublanguage.
To detail OWLOOP, we introduce the implementation of OWL-based \emph{entities} (in Section~\ref{sec:OWLentities}), \emph{expressions} (in Section~\ref{sec:OWLexpressions}) and \emph{axioms} (in Section~\ref{sec:OWLaxioms}) as defined in the OWL-DL formalism.

This Section also presents a notation specifically designed to clarify the differences among the types of symbols involved in an ontology and identify the concepts belonging to the OWL and OOP domains, which are not appreciable with the standard OWL notation. 
Table~\ref{tab:glossary} summarises our notation, which is compliant with the standardised OWL formalism and has been designed to highlight our contribution and simplify its presentation. 
Since our approach addresses general-purpose ontologies, we do not use OWL-based human-readable notations except for a guiding example discussed throughout the paper. 
To improve the readability of the paper, we also introduce acronyms for the standardised OWL expression, which are summarised in Table~\ref{tab:acronyms} for simplicity.

\subsection{OWL Entities}
\label{sec:OWLentities}
\begin{figure}
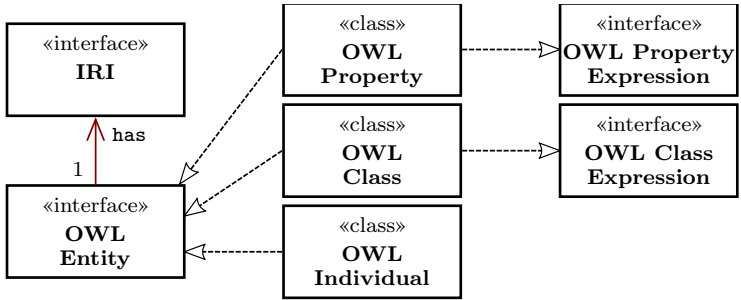
%
    \FIGowlentity%
    \caption{The UML structure of OWL entities and expressions in an ontology.}%
    \label{fig:OWLentity}%
\end{figure}

An ontology represents knowledge through three types of OWL \emph{entities}, \ie classes, properties and individuals.
We derived from \cite{b2012OWL} the Unified Modeling Language (UML) diagram in Figure~\ref{fig:OWLentity} (as well as the others in this Section, \ie Figures~\ref{fig:OWLpropertyAxiom1}--\ref{fig:owlUml2}), which shows through dashed black arrows the specific realisations of the \emph{OWL entity} interface.
Classes and properties in OWL implement the related \emph{expression} interfaces, whereas individuals are related to \emph{assertions}; both aspects are addressed in the next Section.
Furthermore, each entity has an attribute representing an \emph{Internationalised Resource Identifier} (IRI), shown by a red arrow in the diagram.

Entities constitute the vocabulary of an ontology and, we refer to a function \IRIdef that returns an entity given its IRI. 
For clarity, we discriminate between IRIs representing classes with upper-case words, properties with camel-case words and individuals using capitalised names.
The OWL formalism define entities with the following semantics.
\begin{itemize}
    \item We denote an OWL \emph{class} (also referred to as \emph{concept} in the DL formalism) with an upper-case Greek letter, \eg \IRIentity{\Delta}{ROOM}.
          A class collects individuals and is related to other classes via properties. 
    \item We identify an OWL \emph{individual} (or \emph{instance}) with a lower-case Greek letter, \eg \IRIentity{\alpha}{Room1}.
          An individual can be an instance of some classes, and it can be related to other individuals via properties. 
    \item An OWL \emph{property} (or \emph{role}) is described through a bold upper-case Roman letter, \eg \IRIentity{\mathbf{P}}{hasDoor}.
          It spans between the instances of the \emph{domain} and \emph{range} classes (\eg \ONTO{ROOM} and \ONTO{DOOR}), which are related through the property.
\end{itemize}

The OWL formalism introduces \emph{concrete} DL concepts that represent data ranges, \eg numbers, Boolean values, or strings of text.
Their realisations are special individuals called \emph{literals} that can be related to other individuals through \emph{data properties} having a range spanning on concrete classes only.
A data property extends the definition of the $\mathbf{P}$ property above, which is called \emph{object} property in the OWL formalism.
Without any loss of generality, we consider a generic $\mathbf{P}$ to describe both OWL data and object properties.

\subsection{OWL Expressions}
\label{sec:OWLexpressions}
An ontology comprises a \emph{Terminological Box} (TBox), a \emph{Role Box} (RBox) and an \emph{Assertion Box} (ABox).
The TBox can be represented as a hierarchical set of classes (as shown in Figure~\ref{fig:owloop:example}), where arrows represent logic implications.
In the class hierarchy, the OWL formalism defines the $\top$ and $\bot$ concepts as the universal class (\ie \ONTO{THING}) and the empty class (\ie \ONTO{NOTHING}), respectively.
Analogously, the RBox describes a hierarchy of properties.
Instead, the ABox contains a list of individuals represented on the basis of the TBox and RBox, respectively, as realizations of classes (\eg dashed line in Figure~\ref{fig:owloop:example}) and as related with properties (\eg grey arrows in Figure~\ref{fig:owloop:example}).

Each box is defined by a set of axioms encoding knowledge through specified \emph{expressions} \OWL{E} that are syntactic operators applied between entities.
The expression defines the semantics of an axiom and, consequently, the types of entities involved in the axiom.
In particular, axioms in the RBox concern \emph{property expression} $\Expr{\mathbf{P}}$, which describes a property $\mathbf{P}$.
The axioms in the TBox are made of \emph{class expressions} $\Expr{\Delta}$, whereas the ABox contains \emph{assertions} $\Expr{\alpha}$, concerning classes $\Delta$ and individuals $\alpha$, respectively.
Table~\ref{tab:owlAxiom} shows the OWL axioms defined in each box, and it includes \emph{property features}, to define $\mathbf{P}$ with respect to itself, and \emph{class operators}, to define $\Delta$ based on other classes.

\begin{table}%
    \caption{The OWL expressions and the related axioms in the OWL-DL formalism.}%
    \label{tab:owlAxiom}%
    \input{./table/owlAxiom.tex}\end{table}%

\subsection{OWL Axioms}
\label{sec:OWLaxioms}
An axiom is the realisation of a particular expression applied to some entities, and it describes an atomic statement in the ontology. 
An axiom can be considered as a tuple $\OWL{A} = \TUPLE{\OWL{E},E}$, where \OWL{E} is an expression, and $E$ is a set of entities needed to represent the semantics specified by \OWL{E}.
Therefore, each box is an $|n|$-element structure $\{\OWL{A}_1,\ldots,\OWL{A}_n\}$ with expressions \OWL{E} categorised as in Table~\ref{tab:owlAxiom}.
The following Sections detail the axioms and their implementation in the OWL-DL formalism.

\subsubsection{Property Axioms}
\label{sec:propAxioms}
Property axioms are used to represent knowledge in the RBox and involve $\Expr{\mathbf{P}}$ expressions.
The first type of property expression shown in Table~\ref{tab:owlAxiom} is the \emph{Property Subsumption} (PS), which specifies that ${\mathbf{P}\sqsubseteq\mathbf{R}}$, \ie $\mathbf{P}$ implies $\mathbf{R}$.
A PS expression is used to build axioms as
\OWLax{PS}{\mathbf{P},\mathbf{R}},
representing the fact that if two individuals $\alpha$ and $\beta$ are related with $\mathbf{R}$ in the ABox (\ie \ROLE{\alpha}{\beta}{\mathbf{P}}), then $\ROLE{\alpha}{\beta}{\mathbf{R}}$ will also be verified.
It is worth noting that the inverse subsumption operator $\sqsupseteq$ is obtained simply by inverting the elements $\mathbf{R}$ and $\mathbf{P}$.

\begin{figure}
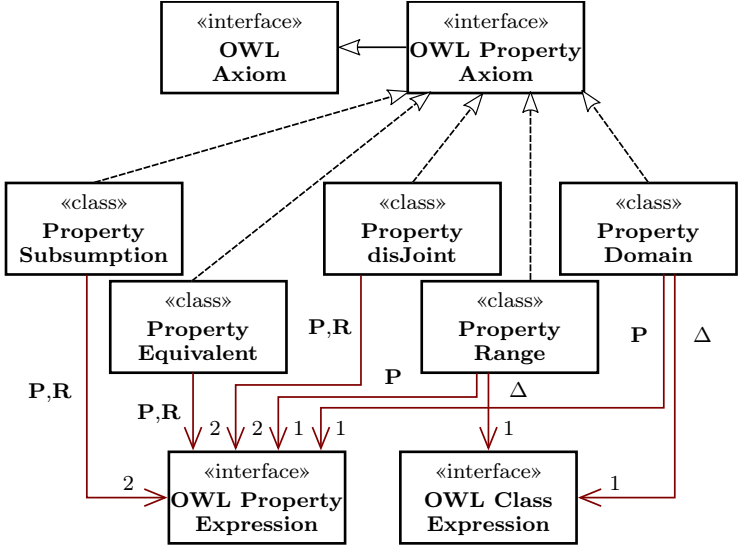
%
    \FIGowlproperty%
    \caption{The UML structure of some OWL property axioms (the others are in Figure~\ref{fig:OWLpropertyAxiom2}).}%
    \label{fig:OWLpropertyAxiom1}%
\end{figure}%

Similarly, the \emph{Property disJoint}, \emph{Property Equivalence}, and \emph{Property Inverse} expressions (PJ, PE and PI, respectively) are used to build axioms in the RBox having ${E\equiv\{\mathbf{P},\mathbf{R}\}}$.
In particular, PJ describes the fact that if \ROLE{\alpha}{\beta}{\mathbf{R}} holds in the ABox, then \ROLE{\alpha}{\beta}{\mathbf{P}} would be inconsistent, \ie ${\mathbf{R}=\neg\mathbf{P}}$.
Furthermore, PE specifies whether $\mathbf{P}$ and $\mathbf{R}$ are equivalent, \ie $\mathbf{R}\equiv\mathbf{P}$,
whereas PI represents that if \ROLE{\alpha}{\beta}{\mathbf{P}} occurs in the ABox, then the axiom \ROLE{\alpha}{\beta}{\mathbf{R}} will always be consistent, such that $\mathbf{P}=\mathbf{R}^{-1}$. 

The \emph{Property Domain} (PD) specifies an axiom \OWLax{PD}{\mathbf{P},\Delta}, which indicates that $\alpha$ must be an instance of the class $\Delta$ for being consistently used in the domain of an axiom involving property $\mathbf{P}$ in the ABox, \ie $\DOMAIN{\mathbf{P}}{\Delta}$.
In this paper, based on the inverse property expression PI, we consider the \emph{Property Range} (PR) defined similarly to PD, but involving the range of a $\mathbf{P}$ realisation in the ABox.

Table~\ref{tab:owlAxiom} also shows property features, \ie an axiom \OWLax{\Expr{\mathbf{P}}}{\mathbf{P}}.
The \emph{Property Functional} (PF) expression represents the fact that a realisation of $\mathbf{P}$ with an individual $\alpha$ as a domain must be unique, \eg in the ABox there can be only one individual for which \IRIentity{\mathbf{P}}{isIn}.
The \emph{Property refleXive} (PX) can involve only the same individual \ONTO{Self}, \ie \ROLE{\alpha}{\alpha}{\mathbf{P}}.
The \emph{Property sYmmetric} (PY) expression specifies that any realisations of the property \ROLE{\alpha}{\beta}{\mathbf{P}} implies \ROLE{\beta}{\alpha}{\mathbf{P}} as well.
The OWL formalism uses PJ-based axioms also to define \emph{asymmetric} properties, which specify that such an implication never occurs.
The \emph{Property Transitive} (PT) indicates that if the realisations \ROLE{\alpha}{\beta}{\mathbf{P}} and \ROLE{\beta}{\gamma}{\mathbf{P}} are listed in the ABox, then also \ROLE{\alpha}{\gamma}{\mathbf{P}} is consistent.
For instance, \ONTO{isConnectedTo} is a property that is symmetric and, possibly, transitive.

\begin{figure}
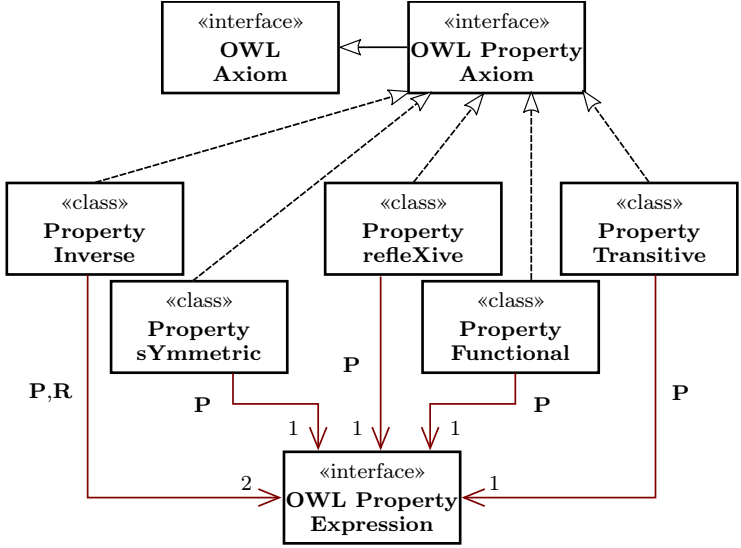
%
    \FIGowlpropertybis%
    \caption{The UML structure of some OWL property axioms (the others are in Figure~\ref{fig:OWLpropertyAxiom1}).}%
    \label{fig:OWLpropertyAxiom2}%
\end{figure}

Figure~\ref{fig:OWLpropertyAxiom1} and Figure \ref{fig:OWLpropertyAxiom2} show the implementation of axioms \OWL{A} in the RBox.
They are \emph{property axioms} that involve \Expr{\mathbf{P}} expressions among entities in $E$, which are shown by the red arrows.
In accordance with Table~\ref{tab:owlAxiom}, Figure~\ref{fig:OWLpropertyAxiom1} shows the PS, PE and PD axioms implemented through $\mathbf{P}$ and $\mathbf{R}$, whereas PD and PR through $\mathbf{P}$ and $\Delta$.
Similarly, Figure~\ref{fig:OWLpropertyAxiom2} shows the PI axiom based on two property expressions, while only one is required for PF, PX, PY and PT.

\subsubsection{Class Axioms}
\label{sec:classAxiom}
The second section of Table~\ref{tab:owlAxiom} is related to axioms describing classes in the TBox through expressions $\Expr{\Delta}$.
The \emph{Class Equivalent} (CE) and \emph{Class disJoint} (CJ) expressions are used to build axioms describing $\Delta$ with respect to another class $\Lambda$, \ie ${E=\{\Delta,\Lambda\}}$.
In particular, the axioms related to CE and CJ specify that if all the individuals classified in the ABox as instances of $\Delta$ are (or are not) realisations of $\Lambda$, then the \emph{vice-versa} holds (or does not hold), \ie ${\Delta\equiv\Lambda}$ and ${\Delta=\neg\Lambda}$, respectively. 

Table~\ref{tab:owlAxiom} also lists an axiom related to the \emph{Class Subsumption} (CS) expression, which involves a set of entities as above, \ie ${E=\{\Delta,\Lambda\}}$.
Axioms based on CS expressions represent individuals of a class that are always instances of another class, but the \emph{vice-versa} does not hold true, \ie ${\ONTO{CORRIDOR}\sqsubseteq\ONTO{LOCATION}}$.
It is noteworthy that such implications can be considered as direct edges in the hierarchy of classes represented in the TBox (\eg arrows in Figure~\ref{fig:owloop:example}), where a CJ expression specifies that no edge can be drawn between two classes, whereas a CE expression results in a bidirectional arrow.

\begin{figure}
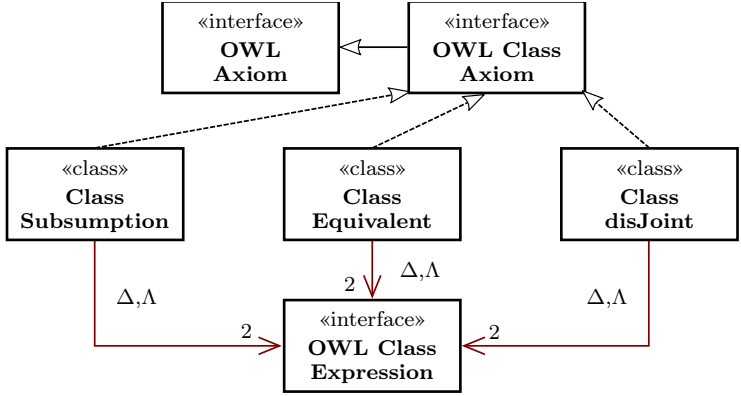
%
    \FIGowlclassaxiom%
    \caption{The UML structure of an OWL class axiom.}%
    \label{fig:OWLclassAxiom}%
\end{figure}

Figure~\ref{fig:OWLclassAxiom} shows the implementation of the \emph{class axioms} that can be encoded in the Tbox, \ie the axioms \OWL{A} involving the above-mentioned $\Expr{\Delta}$ expressions.
In particular, derived class axioms that are related to the CE, CJ and CS expressions have two attributes representing the elements of $E$ as class expressions, \ie ${E=\{\Delta, \Lambda\}}$. 

Table~\ref{tab:owlAxiom} further includes \emph{Class Declaration} (CD), which is used for implementing class expressions based on operations among classes.
A CD axiom requires two class expressions that can either be two classes ${E=\{\Delta,\Lambda\}}$, or a combination of more classes via operators, \ie 
\emph{Class Intersection} (CI), \emph{Class Union} (CU), \emph{Class some Value of} (CV), \emph{Class all value Of} (CO), \emph{Class min cardinality} (Cm) and \emph{Class Max cardinality} (CM) expressions.
Such a combination of classes is possible since the result of any operators is a class $\Pi$ without an explicit name \OWL{N}, \ie an \emph{anonymous class expression}.
For instance, an $\Expr{\Delta}$ expression representing the declaration ${\Delta\doteq\Lambda\sqcup\Phi}$ is encoded as an axiom 
\OWLax{CD}{\Delta,\Pi}, where $\Pi$ is implicitly defined with the axiom \OWLax{CU}{\Lambda,\Phi}. 

The CV, CO, Cm and CM expressions are used for declaring anonymous classes restricting $\Delta$ to classify the individuals involved in specific properties only.
In particular, CV specifies $\Pi\equiv\exists\DOMAIN{\mathbf{P}}{\Delta}$, which is an anonymous class containing all the individuals involved in \emph{some} properties $\mathbf{P}$ with a range spanning in $\Delta$, \ie $\Pi$ classifies all the individuals $\alpha$ involved in \ROLE{\alpha}{\beta}{\mathbf{P}}, where \INST{\beta}{\Delta}.
Similarly, CO defines an anonymous class $\Pi$, but it contains individuals involved \emph{only} in the realisations of $\mathbf{P}$ in the ABox.
Conversely, Cm and CMV restrict the \emph{minimum} and the \emph{maximum} cardinality $c\in\mathbb{N}^+$, \ie the number of $\mathbf{P}$ realisations that should occur for classifying an individual as an instance of $\Pi$.


The OWL formalism also defines the \emph{class has value} expression, which is based on an anonymous class referring to an instance, \ie $\Delta\doteq\exists\DOMAIN{\mathbf{P}}{\alpha}$.
In this paper, we simplify the discussion about that expression thanks to its duality with the CV and CO expressions.

\begin{figure}
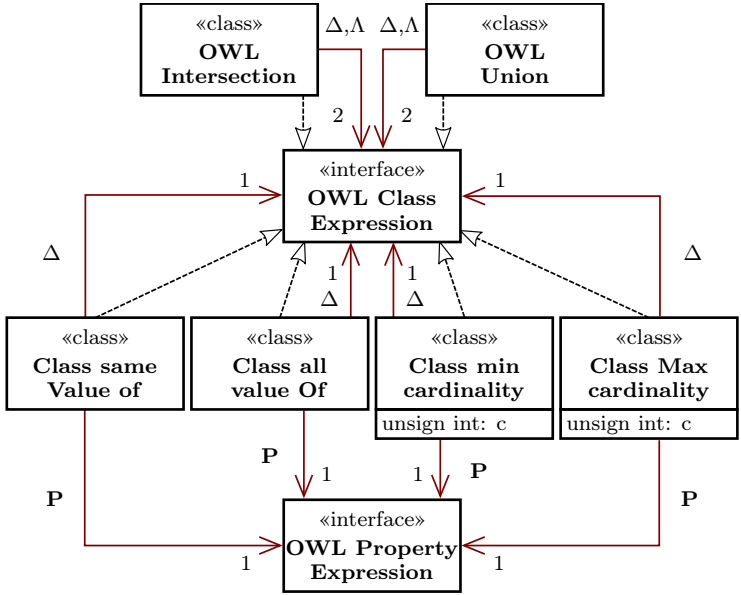
%
    \FIGowlclassexpression%
    \caption{The UML structure of OWL Class Expression composition.}%
    \label{fig:owlUml5}%
\end{figure}

Figure~\ref{fig:owlUml5} shows the implementation of compound class expressions.
In particular, all CI, CU, CV, CO, Cm and CM operators are extensions of class expressions, \ie a named or anonymous class.
They have different attributes as property or class expressions related to OWL entities in accord with Table~\ref{tab:owlAxiom}. 
In particular, CI and CU require two class expressions, \ie $E=\{\Delta,\Lambda\}$, while CV, CO, Cm and CM need a class and a property expression, \ie $E=\{\mathbf{P},\Delta\}$.

\subsubsection{Assertion Axioms}
\label{sec:assertionAxiom}

Assertion axioms are included in the ABox to represent individuals with respect to the knowledge encoded in the TBox or the RBox.
Those axioms are based on $\Expr{\alpha}$ expressions, which are detailed in the third section of Table~\ref{tab:owlAxiom}.
In particular, the \emph{Assertion Class} (AC) expression represents the fact that an individual is an instance of a class, \ie $\alpha{:}\,\Delta$.
The \emph{Assertion Property} (AP) expression represents that $\alpha$ and $\beta$ are related through $\mathbf{P}$, \ie \ROLE{\alpha}{\beta}{\mathbf{P}}.
The \emph{Assertion Same individual} or \emph{Assertion Different individual} expressions (AS and AD, respectively) specify whether two symbols $\alpha$ and $\beta$ refer to the same individual, \ie $\alpha\equiv\beta$, or not, \ie $\alpha=\neg\beta$.

Figure~\ref{fig:owlUml2} shows the expressions derived specifically to represent assertion axioms based on the entities in $E$.
In particular, the AC expression is implemented with attributes $E=\{\alpha, \Delta\}$, while DI and SI with $E=\{\alpha, \beta\}$, and AV with $E=\{\alpha, \beta, \mathbf{P}\}$.

\section{OWLOOP Descriptors}  
\label{sec:OWLOOPDescriptor}
To map OWL axiom into OOP counterparts, OWLOOP exploits interfaces that are shared for all the axioms that might occur in the three boxes of an ontology.
Section~\ref{sec:mapping} defines such a map for each axiom, while this Section presents the OOP interfaces designed to achieve it. 
These interfaces are defined within an OOP hierarchy having the \emph{descriptor} as the base object that implements most of the OWLOOP functionalities.

\subsection{The Internal State of a Descriptor}
Let a \emph{descriptor} \OOP{D} be an OOP \emph{interface} that defines all the methods used to represent a fragment of the ontology through some OWLOOP \emph{axioms}.
The latter are OWL axioms mapped in the OOP domain through \OOP{D}, which requires an OWL ontology \OWL{O}, an expression \OWL{E}, and an \emph{internal state} representing the entities encoded in $E$.
An OOP \emph{class} that implements \OOP{D} can perform OWL to OOP mapping by \emph{reading} OWL axioms from the ontology, and OOP to OWL mapping by \emph{writing} OWLOOP axioms in the ontology.

Although OWL axioms $\OWL{A}_i$ have not been designed with OOP paradigms, a descriptor allows accessing its information in an OOP fashion, \eg with \emph{get} and \emph{set} methods.
Moreover, it is possible to \CODE{build} an element $e \in E$ and obtain another class implementing \OOP{D}, which describes the entity $e$ based on other axioms in the ontology.

More formally, OWLOOP axioms are represented in the internal state of \OOP{D} as a tuple $\OWL{P}=\TUPLE{\OWL{E},x,Y}$, where \OWL{E} is an OWL expression, $x$ is an OWL entity called \emph{ground}, and $Y$ is an |n|-element \emph{entity set} having items $y_i$, that are OWLOOP \emph{entities} (detailed in Section~\ref{sec:mapping}).
Therefore, a descriptor contains $n$ OWLOOP axioms, each mapped to a related OWL axiom $\OWL{A}_i = \TUPLE{\OWL{E}, E}$ with $E=\{x, y_i\}$.

\begin{figure}
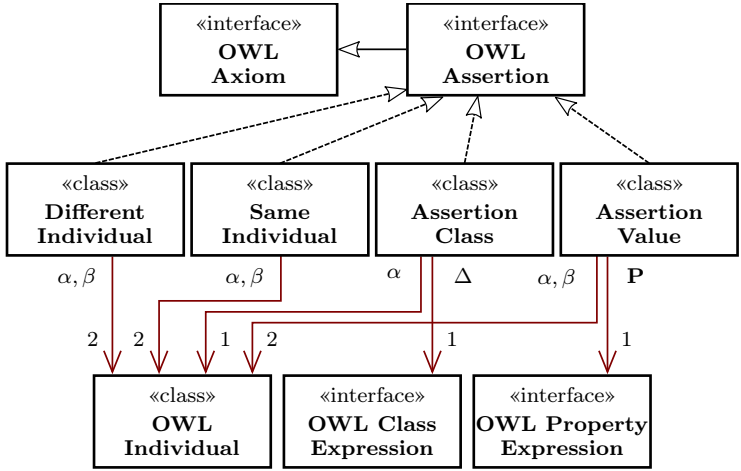
%
    \FIGowlassertion%
    \caption{The UML structure of OWL axiom assertions.}%
    \label{fig:owlUml2}%
\end{figure}%

\subsection{The Interface of a Descriptor}
\label{sec:Dinterface}

The OOP interface shown in Definition~\ref{alg:OWLOOPdescrInterface} presents the methods of a descriptor \OOP{D}.
Each OOP class implementing the \OOP{D} interface must specify an expression \OWL{E}, from which the type of $x$ and $y_i$ are derived using OOP \emph{templates}.
For simplicity, we consider each OOP class implementing \OOP{D} to be \emph{constructed} with empty attributes related to $x$ and $Y$, and with an immutable reference to a software interface \OOP{O} that can manage an ontology \OWL{O}.

Since OWLOOP requires \OOP{O} to be an OOP object, we encapsulated all the components of the OWL-API (which are shown in Figure~\ref{fig:abstract}) in a single object \OOP{O} using the aMOR library\footnote{The aMOR library is available at \url{https://github.com/buoncubi/multi_ontology_reference}.}, which is based on the factory design pattern.
It is worth noticing that \OOP{O} allow accessing low-layer OWL-API functionalities to manipulate and query \OWL{O}, and this assures compatibility with other software that relies on such an API, \eg several reasoners.

Lines~\ref{ln:getEntities}--\ref{ln:setGround} show the methods to access and eventually modify the attributes of \OOP{D}, \ie $x$ and $Y$.
The former is an OWL entity, as defined in Figure \ref{fig:OWLentity}, whereas the latter is a set of OWLOOP entities (defined in Figure~\ref{fig:OWLOOPentity} and presented below).
These \emph{get} and \emph{set} methods are only related to the internal state of \OOP{D} without involving the ontology.

Instead, the \CODE{query} method (defined in Line~\ref{ln:query}) uses the ontology and its reasoner to obtain OWL axioms given \OWL{E} and $x$.
Formally, the \CODE{query} returns all the axioms ${\{\OWL{A}_1, \ldots, \OWL{A}_i,\ldots, \OWL{A}_n\}}$, where ${\OWL{A}_i = \OWLax{\OWL{E}}{x, e_{i1}, \ldots, e_{ij}, \ldots, e_{im}}}$, \ie it retrieves each OWL entity $e_{ij}$ that consistently describes $x$ with the \OWL{E} semantics in the ontology.
A query does not involve OWLOOP axioms and, as a consequence, it does not affect the internal state of \OOP{D}.
Hence, \CODE{query} is a \emph{protected} method, and it should be invoked only within the descriptor implementation, \eg for axioms reading and writing.

The \CODE{read} and \CODE{write} methods (defined in Lines~\ref{ln:read} and \ref{ln:write}) query the ontology and compare the results with the OWLOOP axioms encoded in the internal state of \OOP{D}.
This allows finding the differences between $Y$ and the queried axioms, which are processed differently by the two methods.
In the reading process, $Y$ is changed for becoming equivalent to the queried axioms, while with the writing process, the ontology is changed for containing the same knowledge encoded in $Y$.
OWLOOP keeps track of all the changes and returns a list of \emph{intents} \OOP{I}, which can be useful for further processing, \eg to recover from an inconsistent state of the ontology.

\input{./algorithm/Dinterface.tex}

Line~\ref{ln:build} defines the \CODE{build} method of \OOP{D}, which returns a set of new descriptors with a consistent expression \OWL{E} and a ground $x$ equivalent to the $y_i$ entities of \OOP{D}.
The \CODE{build} method allows accessing the OWL axioms that further describe $y_i$ based on $x$ in an OOP manner.
For example, a descriptor that maps OWL axioms in the ABox involving $x = \ONTO{Robot}$ and $Y = \{\ONTO{Room1}\}$ (\ie \ROLE{x}{y_1}{\ONTO{isIn}}) can build another descriptor grounded in $y_1$ with an entity set encoding its classification in the TBox, \eg axioms expressing that $y_1$ is a \ONTO{ROOM} and a \ONTO{LOCATION}.

The \CODE{build} method relies on the \CODE{newEntityToBuild} method.
The latter is defined in Line~\ref{ln:newbuild} as protected since it should only be invoked from the \CODE{build} method or its extensions.
The \CODE{newEntityToBuild} method simply constructs and returns a new descriptor with a customizable (but consistent) expression \OWL{E}, a ground equivalent to the entity $y_i$ given as input parameter, and an empty entity set.
The \CODE{build} method invokes the \CODE{read} method for initialising the entity set of the descriptor returned by the \CODE{newEntityToBuild} method for each $y_i$.
Hence, the \CODE{build} method returns $n$ descriptors. 

It is worth noting that \OOP{D} implements the \CODE{query}, \CODE{read}, \CODE{write}, and \CODE{build} methods used for all the OWL axioms occurring in the ontology \OWL{O}.
Any OOP classes implementing \OOP{D} can inherit these methods by \emph{overriding} 
($i$)~Lines~\ref{ln:getontology}--\ref{ln:setGround} to provide an ontology interface \OOP{O}, $x$, and $Y$, and 
($ii$)~Line~\ref{ln:newbuild} to construct new descriptors during the building process, \eg as shown in Example~\ref{alg:exDes2}, which is discussed in Section~\ref{sec:examples}.

\begin{table}%
    \caption{The mapping of OWL and OWLOOP axioms \TUPLE{\OWL{E},x,y_i}.}%
    \label{tab:owloopAxiom}%
    \input{./table/owloopAxiom.tex}\end{table}%

\section{OWLOOP to OWL Axioms Mapping}
\label{sec:mapping}

This Section defines a map between the knowledge encoded in a set of OWL axioms ${\{\OWL{A}_1, \OWL{A}_2, \ldots\}}$, where a generic element is ${\OWL{A}_i=\TUPLE{\OWL{E}, E}}$, and the corresponding set of OWLOOP Axioms $\TUPLE{\OWL{E}, x, Y}$, with elements ${\OWL{P}_i=\TUPLE{\OWL{E}, x, y_i}}$.
Our objective is to find a bijective function that transforms $E$ into $Y$.
Since \OOP{D} specifies \OWL{E} and $x$, our map can generically be reduced to $Y = E\setminus\{x\}$.
Sections~\ref{sec:OWLOOPpropAxiom}, \ref{sec:OWLOOPclassAxiom} and \ref{sec:OWLOOPassertionAxiom} detail the map for $\Expr{\mathbf{P}}$, $\Expr{\Delta}$ and $\Expr{\alpha}$ expressions, respectively, while Section~\ref{sec:axiomBuilding} discusses the \CODE{build} method of \OOP{D}.

Our OWL to OOP map can be derived from a comparison of the OWL and OWLOOP axioms shown in Tables~\ref{tab:owlAxiom} and \ref{tab:owloopAxiom}, respectively.
In particular, Table~\ref{tab:owlAxiom} highlights that most of the OWL axioms involve two OWL entities of the same type, \eg PE requires $E=\{\mathbf{P},\mathbf{R}\}$, whereas \OWLax{CS}{\Delta,\Lambda}.
Nevertheless, there are some exceptions.
In particular, axioms concerning property features (\ie PF, PX, PY and PT, presented in Figure~\ref{fig:OWLpropertyAxiom2}) involve one entity only, \ie ${E=\{\mathbf{P}\}}$.
In addition, the axioms related to class operators (\ie PD, PR, CV, CO, Cm and CM, accordingly to Figure~\ref{fig:OWLpropertyAxiom1}) might require a property and a class, \ie ${E=\{\mathbf{P},\Delta\}}$, while AC involves an individual and a class, \ie ${E=\{\alpha,\Delta\}}$, and AV two individuals and a property, \ie ${E=\{\alpha,\beta,\mathbf{P}\}}$ (as shown in Figure~\ref{fig:owlUml2}).
Since \OOP{D} encompasses all OWL axioms, we designed $y_i$ as an OWLOOP entity to take into account these different cases as defined in Figure~\ref{fig:OWLOOPentity} and detailed below.

\subsection{OWLOOP Property Axioms}
\label{sec:OWLOOPpropAxiom}

An OWL property axiom (represented in the RBox through expressions $\Expr{\mathbf{P}}$ and entities $E$ as shown in Figures~\ref{fig:OWLpropertyAxiom1} and \ref{fig:OWLpropertyAxiom2}) is mapped in/from an OWLOOP property axiom ${\OWL{P}_i=\TUPLE{\Expr{\mathbf{P}}, x, Y}}$, where $x=\mathbf{P}$ and the entity set $Y$ encodes the other elements of $E$.

\begin{figure}
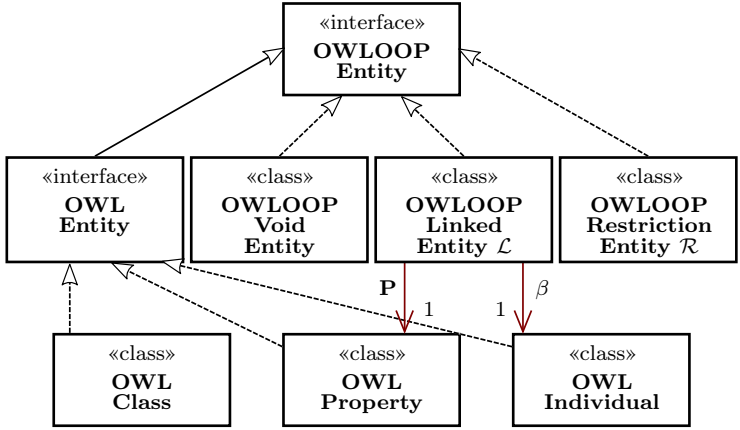
%
    \FIGowloopentity%
    \caption{The UML representing OWLOOP entities derived from OWL entities.}
    \label{fig:OWLOOPentity}%
\end{figure}

Section~\ref{sec:propAxioms} presents property features with ${E=\{\mathbf{P}\}}$, \ie the PF, PY, PX, and PT expressions require the ground entity only.
Therefore, we consider ${Y = E\setminus\{\mathbf{P}\}=\emptyset}$, \ie a \emph{void} structure identified in Figure~\ref{fig:OWLOOPentity} as the OWLOOP \emph{Void Entity}. 

Furthermore, PS, PE, PJ, and PI are expressions defining OWL axioms with ${E=\{\mathbf{P},\mathbf{R}\}}$.
Thus, OWLOOP axioms having ${x=\mathbf{P}}$ are described with an OWLOOP entity ${y_i=\mathbf{R}}$.
The latter is an OWL Property, \ie an OWL Entity, as shown in Figure~\ref{fig:OWLOOPentity}.

If PR and PD expressions define OWL axioms with ${E=\{\mathbf{P},\Delta\}}$, then the relative OWLOOP axioms would be characterised by ${y_i=\Delta}$.
However, since an OWL class expression might also involve anonymous classes (as discussed in Section~\ref{sec:classAxiom}), $y_i$ might become a structure \OOP{R} encoding OWL classes, properties and operators, \ie an OWLOOP \emph{Restriction Entity} (RE), as defined in Figures~\ref{fig:OWLOOPentity} and \ref{fig:owloopRestriction}.
In accordance with Table~\ref{tab:owloopAxiom}, $y_i$ for PR and PD is generically based on \OOP{R}, and the next Section details its definition.

\subsection{OWLOOP Class Axioms}
\label{sec:OWLOOPclassAxiom}
An OWL class axiom encoded in the TBox through expressions $\Expr{\Delta}$ and entities $E$ (shown in Figures~\ref{fig:OWLclassAxiom} and \ref{fig:owlUml5}) is mapped to an OWLOOP class axiom ${\OWL{P}_i = \TUPLE{\OWL{E}_{\mathbf{\Delta}},\Delta,y_i}}$, where $y_i$ is related to the elements of $E$ except for the ground entity $\Delta$.

Table~\ref{tab:owlAxiom} shows that axioms based on the CS, CE and CJ expressions involve entities $E = \{\Delta,\Lambda\}$.
Therefore, the OWLOOP entities for these axioms are OWL classes $y_i=\Lambda$.

As discussed above for PD and PR expressions, also CD-based axioms can be mapped as $y_i = \Lambda$ if no anonymous classes are considered.
Otherwise, we need to consider a more general $|n|$-element entity set $Y = \{\OOP{R}_1, \ldots, \OOP{R}_n\}$, accordingly with Table~\ref{tab:owloopAxiom}. 

\begin{figure}
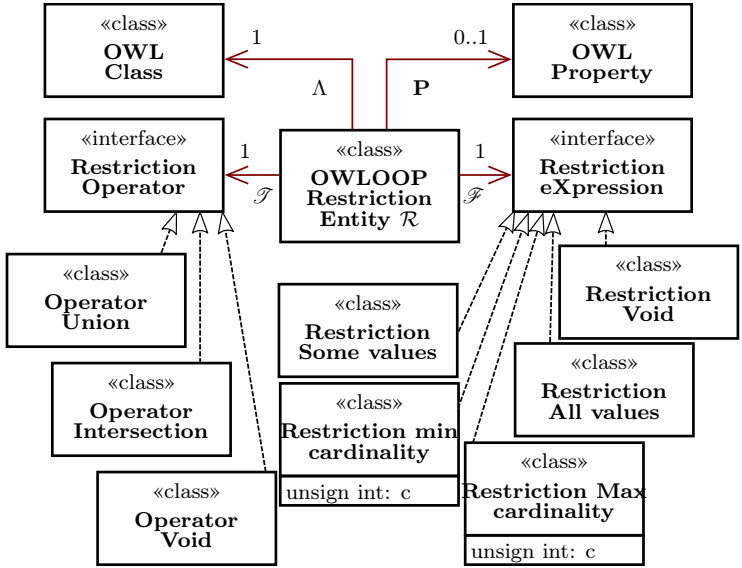
%
    \FIGowlooprestriction
    \caption{The UML structure of the OWLOOP restriction \OOP{R} for mapping the compound OWL class expressions shown in Figure~\ref{fig:owlUml5}.}%
    \label{fig:owloopRestriction}%
\end{figure}

Figure~\ref{fig:owloopRestriction} shows the RE structure, which is a tuple ${\OOP{R}_i = \TUPLE{\OWL{T},\OWL{F},Z}}$, where \OWL{T} is a \emph{Restriction Operator} (RO), \OWL{F} indicates a \emph{Restriction eXpression} (RX), and $Z$ contains OWL entities and parameters required to represent the related OWL expression \OWL{E}.
In particular, \OWL{T} is used for mapping the CU and CI OWL expressions (shown in Figure~\ref{fig:owlUml5}) with a value identifying the operator related to the $\OOP{R}_{i+1}$ element in $Y$.
Therefore, \OWL{T} can be an \emph{Operator Union}, \emph{Intersection} or \emph{Void} (OU, OI and OV, respectively), whereby the latter is used if $\OOP{R}_{i+1}$ does not exist.
The type of the restriction \OOP{R} is denoted by \OWL{F}, which spans among the OWLOOP expressions related to the CV, CO, Cm, and CM expressions, respectively:  \emph{Restriction Some value} (RS), \emph{Restriction All values} (RA), and \emph{Restriction min} or \emph{Max cardinality} (Rm and RM).
Figure~\ref{fig:owloopRestriction} also introduces the \emph{Restriction Void} (RV) to represent the simple cases where \OOP{R} only involves another OWL entity of the same type of the ground entity. 
IF \OWL{F} is not RV, $Z$ either requires a class $\Lambda$ or a property $\mathbf{P}$ with a range $\Lambda$ and, eventually, a cardinality $c$.

For example, let us consider OWL axioms defining a class using intersections of named and anonymous classes, \eg ${\Delta\doteq\Lambda \sqcap \exists\mathbf{P.}\Phi \sqcap {\geqslant}2\,\mathbf{P.}\Phi}$, where ${\IRIentity{\Delta}{CORRIDOR}}$, \IRIentity{\Lambda}{LOCATION}, \IRIentity{\mathbf{P}}{hasDoor} and ${\IRIentity{\Phi}{DOOR}}$.
In this case, we could implement a descriptor with the ground $\Delta$ and the entity set $Y = \{\OOP{R}_1, \OOP{R}_2, \OOP{R}_3\}$ for representing the related OWLOOP axioms.
In particular, $\OOP{R}_1 = \OWLax{RI, RV}{\Lambda}$ represents that a corridor is a type of location.
This is in conjunction with $\OOP{R}_2 = \OWLax{RI, RS}{\mathbf{P}, \Phi}$, which describes that a corridor should have some doors.
The latter is in conjunction with $\OOP{R}_3 = \OWLax{RV, Rm}{2, \mathbf{P}, \Phi}$, which restricts the corridor to have at least two doors.

It is worth noticing that, if $Y$ involves \OOP{R} structures, then $Y$  should be ordered for preserving the operation precedences among the axioms that define a class.
Moreover, in the case of complex definitions involving parenthesis, the definition of \OWL{T} can be extended to accommodate a tree-like representation, \eg an AST.
Such an extension would be similar to the approach based on anonymous classes exploited by OWL-API, but it would lead to issues regarding the \CODE{build} method of \OOP{D}, which are discussed in Section~\ref{sec:axiomBuilding}.

\subsection{OWLOOP Assertion Axioms}
\label{sec:OWLOOPassertionAxiom}

An OWL assertion axiom encoded in the ABox through expressions $\Expr{\alpha}$ and entities $E$ (as shown in Figure~\ref{fig:owlUml2}) is mapped to an OWLOOP assertion axiom ${\OWL{P}_i = \TUPLE{\OWL{E}_{\mathbf{\alpha}},\alpha,y_i}}$, where $y_i$ is related to the elements of $E$ except for the ground entity $\alpha$.

Section~\ref{sec:assertionAxiom} discussed that axioms based on the AS and AD expressions require ${E = \{\alpha,\beta\}}$, therefore ${x = \alpha}$ and ${y_i = \beta}$.
In addition, AC-based axioms involve ${E = \{\alpha,\Delta\}}$, and therefore ${y_i=\Delta}$; in accordance with Table~\ref{tab:owloopAxiom}.

Table~\ref{tab:owlAxiom} shows that AV-based axioms require $E = \{\alpha,\beta,\mathbf{P}\}$.
Hence, we define another type of OWLOOP entity (as shown in Figure~\ref{fig:OWLOOPentity}), which is the \emph{Linked Entity} (LE) ${\OOP{L} = \{\mathbf{P},\beta\}}$, and we pose $y_i = \OOP{L}_i$.
For instance, the OWL axioms \ROLE{\alpha}{\beta}{\mathbf{P}} and \ROLE{\alpha}{\gamma}{\mathbf{R}} are mapped into OWLOOP axioms as
\TUPLE{\text{AV},\alpha,\{\OOP{L}_1,\OOP{L}_2\}}, where $\OOP{L}_1{=}\{\mathbf{P},\beta\}$ and $\OOP{L}_2{=}\{\mathbf{R},\gamma\}$.

We grounded the AC-based axioms (\ie \INST{\alpha}{\Delta}) in $\alpha$ to represent in $Y$ the classes in which it is an instance of, \ie $\Delta$.
Nevertheless, we could also ground such an axiom in $\Delta$ to represent the individuals that are its realization in $Y$.
With the latter viewpoint, we can define another OWLOOP class axiom that would be mapped into the AC-based OWL axiom, and we call it \emph{Class Assertion} (CA), \ie \TUPLE{\text{CA},\Delta,\alpha}.
A similar consideration can also be done for the OWLOOP axioms encoding more than one OWL entity in $y_i$.

\subsection{OWLOOP Axiom Building}
\label{sec:axiomBuilding}

Section~\ref{sec:Dinterface} introduced the \CODE{build} method defined in the descriptor interface \OOP{D}.
Given an OOP object implementing \OOP{D}, \ie $\OOP{M}^{\OOP{D}(\OWL{E})}$, which internally encodes OWLOOP axioms with the $y_i^\OOP{M}$ entities, the \CODE{build} method returns a set of descriptors $\OOP{N}^{\OOP{D}(\bar{\OWL{E}})}$ concerning a consistent expression $\bar{\OWL{E}}$ and having the ground set as $x^\OOP{N} = y_i^\OOP{M}$.

The building operation is straightforward when the entity set of $\OOP{M}^{\OOP{D}(\OWL{E})}$ contains elements $y_i^\OOP{M}$ that are OWL entities.
However, if $y_i^\OOP{M}$ are structures like \OOP{R} or \OOP{L}, the \CODE{build} method might become badly defined.
Nonetheless, since \OOP{R} and \OOP{L} contain OWL entities, it is always possible to specify a building method based on their structures, but at the price of losing a certain degree of generality.

In particular, an object implementing the AV-based descriptor interface $\OOP{M}^{\OOP{D}(\text{AV})}$ involves entities $y_i$ that are $\OOP{L}_i$ structures, \ie pairs $\{\mathbf{P},\beta\}$.
Thus, the \CODE{build} method can return two types of descriptors: one involves \Expr{\alpha} expressions with ground ${x^{\OOP{N}} = \beta}$, while the other concerns \Expr{\mathbf{P}} expressions with ground ${x^{\OOP{N}} = \mathbf{P}}$.
We consider the default building policy for $\OOP{D}(\text{AV})$ (defined in Line~\ref{ln:build}) returning individual descriptors grounded on $\beta$.
In addition, we extend $\OOP{D}(\text{AV})$ with a \CODE{buildProperty} method that returns property descriptors grounded on $\mathbf{P}$.
Consequently, we also introduce a \CODE{newPropertyToBuild} method in $\OOP{D}(\text{AV})$, which requires a property $\mathbf{P}$ as input and returns a new descriptor with an empty internal state concerning \Expr{\mathbf{P}} expressions, similarly to Line~\ref{ln:newbuild}.

Since the structure of \OOP{R} is unknown \emph{a priori}, designing a general building process is far from trivial if \OOP{R} occurs in the entity set of a descriptor.
Nonetheless, if the knowledge encoded in an ontology is known, it would always be possible to design building approaches that identify grounding entities within \OOP{R} structures because they encode OWL entities.
This paper assumes that the \CODE{build} method of the $\OOP{D}(\text{PR})$, $\OOP{D}(\text{PD})$ and $\OOP{D}(\text{CD})$ descriptors is only defined if they involve restrictions \OOP{R} having the void OWLOOP expression and operator, \ie \OOP{R} is reduced to an OWL entity, accordingly to Section~\ref{sec:OWLOOPclassAxiom}.

Finally, the building methods of descriptors related to property features, \ie PF, PY, PX and PT, are undefined since their entity set is a void structure and, as such, $\nexists y_i$.

\section{Implementation of OWLOOP Descriptors}
\label{sec:implementation}

This Section addresses the implementation of OOP classes that realise the descriptor interface \OOP{D} detailed in Section~\ref{sec:OWLOOPDescriptor}.
With reference to Table~\ref{tab:owloopAxiom}, Section~\ref{sec:OWLOOPDescriptorComposition} introduces \emph{compound} descriptors, which implement some descriptor interfaces to encompass \Expr{\mathbf{P}}, \Expr{\Delta} and \Expr{\alpha} expressions, addressed in Sections~\ref{sec:propDescriptor}, \ref{sec:fullClassDescriptor} and \ref{sec:fullIndividual}, respectively.

\subsection{Composition of Descriptors}
\label{sec:OWLOOPDescriptorComposition}

Section~\ref{sec:mapping} derives three extensions of \OOP{D} by specifying its template parameter \Expr{}, \ie
  ($i$)~the \emph{property descriptor} ${\OOP{D}(\Expr{\mathbf{P}})}$ has a property as ground, \ie ${x = \mathbf{P}}$, and it describes axioms based on a property expression \Expr{\mathbf{P}}, which is one item of ${P = \{\text{PS},}$ PJ, PE, PI, PD, PR, PF, PX, PY, PT\},
 ($ii$)~the \emph{class descriptor} ${\OOP{D}(\Expr{\Delta})}$ has $x = \Delta$, and \Expr{\Delta} spans in ${C = \{\text{CE}}$, CJ, CS, CD, CA\}, and
($iii$)~the \emph{individual descriptor} ${\OOP{D}(\Expr{\alpha})}$ has $x = \alpha$, and \Expr{\alpha} spans in ${A = \{\text{AC}}$, AV, AS, AD\}.

We define a \emph{compound descriptor} as an OOP class \OOP{K} implementing some descriptor interfaces $\OOP{K}^{\OOP{D}(\Expr{1}), \ldots, \OOP{D}(\Expr{d})}$ (that we might simplify to $\OOP{K}^{\Expr{1}, \ldots, \Expr{d}}$),which encodes the structure $\{\TUPLE{\Expr{1},x,Y_1}, \ldots, \TUPLE{\Expr{i},x,Y_i}, \ldots, \TUPLE{\Expr{d},x,Y_d}\}$.
Hence, \OOP{K} implements a \emph{property descriptor} when $x$ is a property, and \OOP{K} involves only \Expr{\mathbf{P}} expressions, \ie each $\Expr{i}\in P$.
Similarly, \OOP{K} realises a \emph{class descriptor} when $x$ is a class and each $\Expr{i}\in C$, whereas \OOP{K} implements an \emph{individual descriptor} if $x$ is an individual and each $\Expr{i}\in A$.

In this way, \OOP{K} collects different types of axioms describing an OWL entity $x$.
In particular, the immutable references of \OOP{D}, \ie the ontology interface \OOP{O} and the expression \OWL{E} (defined in Lines~\ref{ln:expression} and \ref{ln:getontology} of Definition~\ref{alg:OWLOOPdescrInterface}), as well as the ground $x$ (Lines~\ref{ln:getGround} and \ref{ln:setGround}), are shared among all the $\OOP{D}(\Expr{i})$ realisations of \OOP{K}.
Moreover, the entity set and the building operations (Lines~\ref{ln:getEntities}, \ref{ln:newbuild} and \ref{ln:build}) are specialised for each $\OOP{D}(\Expr{i})$.
Instead, the querying, reading and writing methods (Lines~\ref{ln:query}, \ref{ln:read} and \ref{ln:write}) are implemented using a sequential call to the methods inherited from all \Expr{i}.

For instance, let us consider a class descriptor $\OOP{K}^{\text{CA},\text{CJ}}$ having a ground entity $x = \Delta$ accessible through the $\OOP{K}\CODE{.getGround}()$ method, which returns the same $x$ for both \OOP{D} realisations.
Instead, \OOP{K} would construct two distinct entity sets (\ie $Y^\text{CA}$ and $Y^\text{CJ}$), which can be accessed with the $\OOP{K}\CODE{.CA.getEntities}()$ and $\OOP{K}\CODE{.CJ.getEntities}()$ methods.
Similarly, the two building processes can be invoked through $\OOP{K}\CODE{.CA.build}()$ and $\OOP{K}\CODE{.CJ.build}()$, which also involve the respective \CODE{newEntityToBuild} methods.
Differently, the method $\OOP{K}\CODE{.read}()$ encapsulates the $\OOP{K}\CODE{.CA.read}()$ and $\OOP{K}\CODE{.CJ.read}()$ methods by calling them sequentially, and this also holds for $\OOP{K}\CODE{.write}()$.
Example~\ref{alg:exDesF}, which is discussed in Section~\ref{sec:examples}, shows another example regarding the composition of descriptors.

As introduced, for each $\OOP{D}(\Expr{i})$ interface that \OOP{K} realises, \OOP{K} needs to specify the descriptors that would be built (\ie the type of descriptor that the \CODE{newEntityToBuild} method returns), which must be another OOP class implementing some descriptor interfaces $\OOP{Q}^{\Expr{1}\ldots\Expr{q}}$ with a consistent ground.
In other words, if an entity set of \OOP{K} to build is made up of properties, $\OOP{Q}$ would rely only on property descriptors.
Instead, if the entity set is made of classes or individuals, then \OOP{Q} must implement class or individual descriptors, respectively.
It is worth noticing that if \OOP{Q} implements different combinations of the \Expr{i} expressions, then \OOP{K} would build new descriptors \OOP{Q} representing different types of axioms.

\begin{figure}
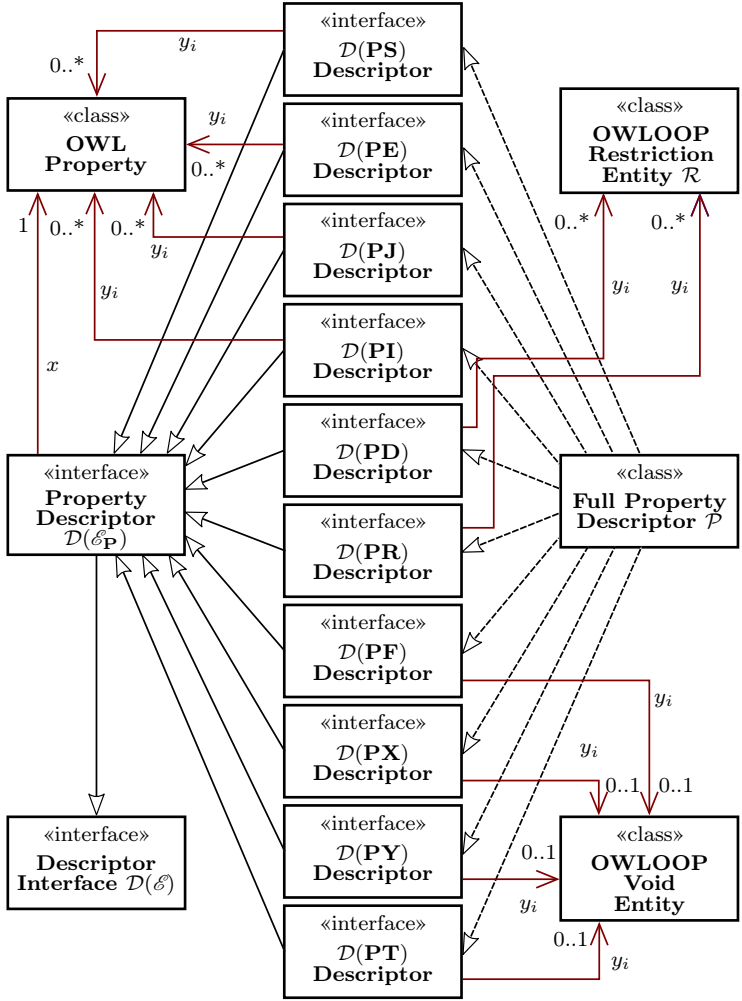
%
    \FIGdescriptorproperty%
    \caption{The implementation of the \emph{full property descriptor} \OOP{P}.}%
    \label{fig:OWLOOPdescrProperty}%
\end{figure}%

\subsection{Compound Property Descriptors}
\label{sec:propDescriptor}

This Section addresses the \emph{Full Property Descriptor} $\OOP{P}^{\Expr{1}\ldots\Expr{q}}$, \ie an OOP class that realises the $\OOP{D}(\Expr{i})$ interface for each expression $\Expr{i}\in P$.
It is worth noticing that any definable property descriptor is a simplified case of \OOP{P} since it involves a subset of $P$.

OWLOOP provides an OOP interface extending \OOP{D} for all the \Expr{\mathbf{P}} expressions, which are collected in the set $P$ and shown in the first part of Table~\ref{tab:owloopAxiom}, \ie the \emph{Property}
\emph{Subsumption} $\OOP{D}(\text{PS})$, 
\emph{Equivalence} $\OOP{D}(\text{PE})$, 
\emph{disJoint} $\OOP{D}(\text{PJ})$, 
\emph{Inverse} $\OOP{D}(\text{PI})$, 
\emph{Range} $\OOP{D}(\text{PR})$, 
\emph{Domain} $\OOP{D}(\text{PD})$, 
\emph{Functional} $\OOP{D}(\text{PF})$, 
\emph{sYmmetric} $\OOP{D}(\text{PY})$, 
\emph{refleXive} $\OOP{D}(\text{PX})$, and
\emph{Transitive} $\OOP{D}(\text{PT})$ \emph{Descriptor}.
Compound property descriptors must realise combinations of these interfaces, and \OOP{P} encompasses all of them, as shown in Figure~\ref{fig:OWLOOPdescrProperty}.

The UML shown in Figure~\ref{fig:OWLOOPdescrProperty} defines \OOP{P} with a ground that is shared with all $\OOP{D}(\Expr{\mathbf{P}})$, while the entity set changes on the basis of the related expression.
In particular, descriptors concerning PF, PY, PX and PT expressions (\ie property features) have a void entity set, whereas descriptors addressing PS, PE, PJ and PI have an entity set made of OWL properties, and the entities encoded for PR and PD are restrictions \OOP{R} involving OWL property and classes.
In accordance with Section~\ref{sec:axiomBuilding}, we define the \CODE{build} methods for PS, PE, PJ and PI to return other property descriptors implementing some $\OOP{D}(\Expr{\mathbf{P}})$.

\begin{figure}
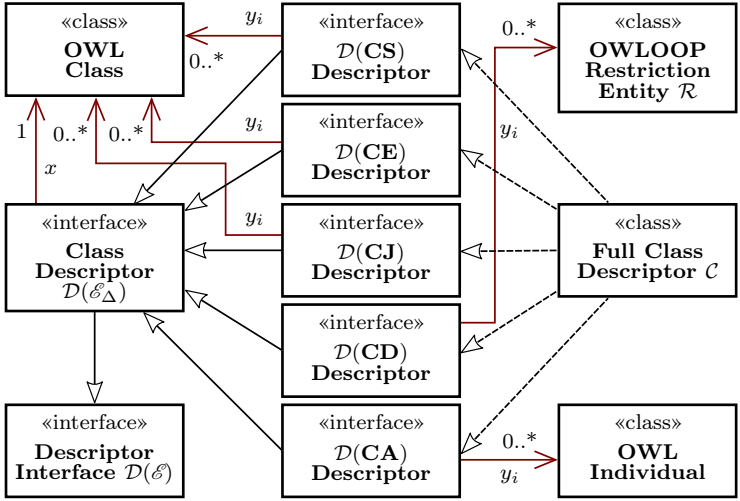
%
    \FIGdescriptorclass    
    \caption{The implementation of the \emph{full class descriptor} \OOP{C}.}
    \label{fig:OWLOOPdescrClass}%
\end{figure}%

\subsection{Compound Class Descriptors}
\label{sec:fullClassDescriptor}

This Section presents the \emph{Full Class Descriptor} $\OOP{C}^{\Expr{1}\ldots\Expr{q}}$, which concerns all the $\Expr{i}\in C$.
Similarly to above, any class descriptor based on a subset of $C$ is a simplified case of \OOP{C}.

OWLOOP provides extensions of \OOP{D} related to each class expression $\Expr{\Delta}$, which are collected in $C$.
In particular, those interfaces are the \emph{Class}
\emph{Declaration} $\OOP{D}(\text{CD})$, 
\emph{Subsumption} $\OOP{D}(\text{CS})$,
\emph{Equivalence} $\OOP{D}(\text{CE})$,
\emph{disJoint} $\OOP{D}(\text{CJ})$, and
\emph{Assertion} $\OOP{D}(\text{CA})$ \emph{Descriptor}.

According to Table~\ref{tab:owloopAxiom}, Figure~\ref{fig:OWLOOPdescrClass} shows that the CD-based descriptor is characterised by an entity set made of restrictions \OOP{R}, whereas the entity set of the descriptors based on the CS, CE and CJ expressions are made of OWL classes, and CA has $Y$ made of OWL individuals.
Hence, the \CODE{build} method related to $\OOP{D}(\text{CA})$ creates new individual descriptors $\OOP{D}(\Expr{\alpha})$, while descriptors specialised for the CS, CE, and CJ expressions build class descriptors $\OOP{D}(\Expr{\Delta})$, in accordance with Section~\ref{sec:axiomBuilding}.

\subsection{Compound Individual Descriptor}
\label{sec:fullIndividual}

This Section presents the \emph{Full Individual Descriptor} $\OOP{A}^{\Expr{1}\ldots\Expr{q}}$, which concerns all the $\Expr{i}\in A$.
Any individual descriptor based on a subset of $A$ is a simplified case of \OOP{A}.

\begin{figure}
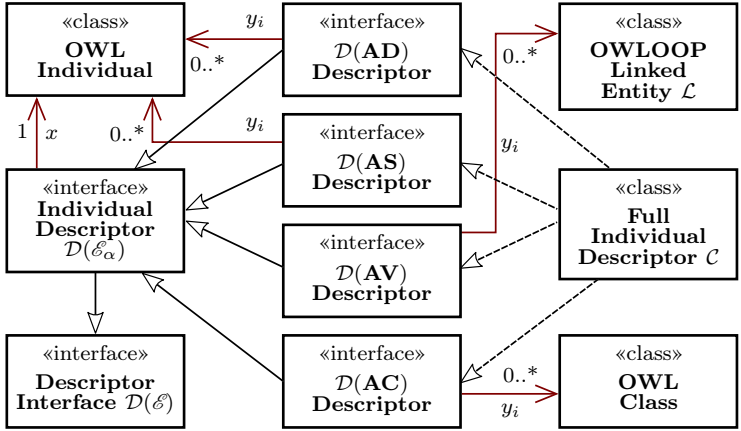
%
    \FIGdescriptorindividual      
    \caption{The implementation of the \emph{full individual descriptor} \OOP{A}.}
    \label{fig:OWLOOPdescrInd}%
\end{figure}

OWLOOP provides an extension of \OOP{D} for each assertion $\Expr{\alpha}$ implemented by \OOP{A}, as shown in Figure~\ref{fig:OWLOOPdescrInd}.
In particular, those interfaces are the \emph{Assertion}
\emph{Class} $\OOP{D}(\text{AC})$, 
\emph{Value} $\OOP{D}(\text{AV})$, 
\emph{Same individual} $\OOP{D}(\text{AS})$, and
\emph{Different individual} $\OOP{D}(\text{AD})$ \emph{Descriptor}.

According to Table~\ref{tab:owloopAxiom}, each interface \OOP{D} that \OOP{A} realises has an entity set $Y$ that changes based on the related expression.
In particular, since $Y^\text{AC}$ is a set of OWL classes, the related \CODE{build} method returns class descriptors $\OOP{D}(\Expr{\Delta})$.
Instead, since $Y^\text{AS}$ and $Y^\text{AD}$ are sets of OWL individuals, the related \CODE{build} methods return individual descriptors $\OOP{D}(\Expr{\alpha})$.

AV-based descriptors involve \OOP{L} structures, which lead to two distinct building methods.
As detailed in Section~\ref{sec:axiomBuilding}, $\OOP{D}(\text{AV})$ might either build new property descriptors $\OOP{D}(\Expr{\mathbf{P}})$ or new individual descriptors $\OOP{D}(\Expr{\alpha})$.

\section{Usage Examples of OWLOOP}
\label{sec:examples}

This Section discusses three use cases based on the scenario sketched in Figure~\ref{fig:owloop:example}, where the goal is to represent a topological map for service robots operating in a smart environment.
In particular, Section~\ref{sec:example1} presents how to use a descriptor to get, set, read and write OWL axioms subjected to reasoning.
Section~\ref{sec:example2} focuses on axiom building, and Section~\ref{sec:compaundExample} addresses compound descriptors and OOP hierarchies. 

\subsection{Reading and Writing using Descriptors}
\label{sec:example1}

Let us implement a function that adds a new \ONTO{LOCATION} in the ontology \OWL{O} depicted in Figure~\ref{fig:owloop:example}.
The function should create OWL axioms to represent a new individual \IRIentity{\alpha}{Location3}, which is connected to \IRIentity{\beta}{Corridor1} through \IRIentity{\gamma}{Door3}.
Example~\ref{alg:ex1} shows such a function, which returns the IRIs of the OWL classes categorising the new location $\alpha$, \ie \ONTO{ROOM} and its parent classes because $\alpha$ has only one door.

Example~\ref{alg:ex1} exploits \OOP{A} (presented in Section~\ref{sec:fullIndividual}) to implement the function, which inputs encompass the ontology interface \OOP{O},  the IRIs related to the new individual \CODE{I$_\alpha$}, the connected \ONTO{LOCATION} \CODE{I$_\beta$} and the related \ONTO{DOOR} \CODE{I$_\gamma$}.
The function also relies on a constant IRI \CODE{I$_\mathbf{P}$} representing \ONTO{LOCATION} having doors through an OWL property ${\IRIentity{\mathbf{P}}{\CODE{I$_\mathbf{P}$}} = \IRI{\ONTO{hasDoor}}}$.

At Lines~\ref{ln:ex1:IRIground1} and \ref{ln:ex1:IRIground2}, input IRIs are used to retrieve the related OWL entities through \OOP{O}.
At Lines~\ref{ln:ex1:LocDescr1}--\ref{ln:ex1:ConDescr2}, two \OOP{A} descriptors are constructed with a ground $x = \alpha$ and $x = \beta$. 
At Lines~\ref{ln:ex1:Prop1} and \ref{ln:ex1:Prop2}, an ${\OOP{L} = \TUPLE{\mathbf{P},\gamma}}$ structure is added to both descriptors for representing that $\alpha$ and $\beta$ share the same door $\gamma$.
The added OWLOOP axioms based on \OOP{L} (\ie \ROLE{\alpha}{\gamma}{\mathbf{P}} and \ROLE{\beta}{\gamma}{\mathbf{P}}), are then written in the ontology at Lines~\ref{ln:ex1:Locwrite} and \ref{ln:ex1:Conwrite}.

Line~\ref{ln:ex1:reason} performs OWL reasoning to update the ontology by inferring new axioms based on the changes we made in the previous two Lines.
Line~\ref{ln:ex1:read} reads the inferred axioms, and the internal state of the descriptor grounded on $\alpha$ changes, \ie the AV-based entity set of \OOP{A} now describes that \ROLE{\alpha}{\beta}{\ONTO{isConnectedTo}}, while $Y^\text{AC}$ encodes that $\alpha$ is an instance of ${\{\top,\ONTO{LOCATION},\ONTO{INDOOR},\ONTO{ROOM}\}}$; which is returned at Line~\ref{ln:ec1:out}.


\input{./algorithm/example1.tex}

\subsection{Building Descriptors and their Definitions}
\label{sec:example2}

In this use case, we want that the robot knowns which types of locations are reachable from its position.
Thus, we design a specific OOP class \OOP{H} implementing $\OOP{D}(\text{CS})$, \ie $\OOP{H}^{{CS}}$, as shown in Line~\ref{ln:ex2:implements} of Example~\ref{alg:exDes2}.
According to Tables~\ref{tab:owlAxiom} and \ref{tab:owloopAxiom}, the CS expression is grounded on an OWL Class and describes its subsuming axioms.
Therefore, \OOP{H} can be used to retrieve the most specific representation of a \ONTO{LOCATION}, \ie a \ONTO{ROOM} or a \ONTO{CORRIDOR}, which are disjoint leaves in the TBox tree, as shown in Figure~\ref{fig:owloop:example}.

Example~\ref{alg:exDes2} details the implementation of \OOP{H}, which is constructed at Line~\ref{ln:ex2:constr} with an ontology interface \OOP{O}, a ground class $x$, and an empty entity set%
    \footnote{We always consider lists without duplications, \ie in the Examples we denote a \emph{set} with $[\,]$, which neither contains equivalent IRIs, nor descriptors with the same ground and expressions.}
$Y^{\text{CS}}$ that will contain subsuming classes.
Since \OOP{H} provides these attributes, \ie Lines~\ref{ln:ex2:getO}--\ref{ln:ex2:getY}, \OOP{H} inherits from \OOP{D} the ability to \CODE{query}, \CODE{read} and \CODE{write}, \ie Lines~\ref{ln:query}--\ref{ln:write} of Definition~\ref{alg:OWLOOPdescrInterface}.

To inherit the \CODE{build} ability, \OOP{H} \emph{overrides} the \CODE{newEntityToBuild} method as shown in Line~\ref{ln:ex2:newBuild}.
The latter would return class descriptors grounded on the items of $Y^{\text{CS}}$.
In this Example, we build other instances \OOP{H} to further describe subsumed classes. 
As described below, such a recursive building allows implementing the checking method \ONTO{isLeaf} conveniently.
Indeed, at Line~\ref{ln:ex2:leaf}, it is sufficient to check whether $Y^{\text{CS}}$ contains only an item that is $\bot$ (\ie \ONTO{NOTHING}) to infer that $x$ is a leaf class in the TBox.

\input{./algorithm/example2D-H.tex}
\input{./algorithm/example2D-F.tex}

To address this use case, we also require the OOP class $\OOP{B}^{\text{AV},\text{AC}}$, which is a compound individual descriptor encompassing $Y^\text{AV}$, \ie a set of \OOP{L} structures, and $Y^\text{AC}$, \ie a set of classes.
Example~\ref{alg:exDesF} shows the implementation of \OOP{B}, which uses the same approach presented to implement \OOP{H} twice, as shown in Lines~\ref{ln:ex3:start}--\ref{ln:ex3:end}.
In accord with Section~\ref{sec:OWLOOPDescriptorComposition}, Lines~\ref{ln:ex3:combine1}--\ref{ln:ex3:combine2} define the extension of the \CODE{read}, \CODE{write} and \CODE{query} methods when a descriptor involves more than one expression; in this case, AV and AC.

With reference to Tables~\ref{tab:owlAxiom} and \ref{tab:owloopAxiom}, the AV expression is mapped into the \ROLE{\alpha}{\beta}{\mathbf{P}} OWL axiom, while AC into \INST{\alpha}{\Delta}, where \OOP{B} has ground $x=\alpha$.
To implement the $\OOP{D}(\text{AC})$ interface, \OOP{B} needs to override the \CODE{newEntityToBuild} method such to return a class descriptor.
At Line~\ref{ln:ex3:end}, we configure \OOP{B} to return a class descriptor representing subsuming axioms, \ie given $\alpha$, \OOP{B} builds descriptor \OOP{H} grounded on $\Delta$.
As discussed in Section~\ref{sec:axiomBuilding}, \OOP{B} implements the $\OOP{D}(\text{AV})$ interface when it overrides the \CODE{newEntityToBuild} and the \CODE{newPropertyToBuild} methods, which should return an individual and a property descriptor, respectively.
At Line~\ref{ln:ex2:newBuild1}, we design \OOP{B} to build other instances of \OOP{B}, \ie given $\alpha$, \OOP{B} builds descriptors with $x=\beta$ and involving AV and CS expressions.
For the sake of completeness, at Line~\ref{ln:ex2:newBuild2}, \OOP{B} also builds instances of \OOP{P}, which is presented in Section~\ref{sec:propDescriptor}.

Line~\ref{ln:ex2DD:buildRangeProp} implements a convenient method to build individuals descriptors from AV-based axioms given an OWL \CODE{property} $\mathbf{P}$.
At Lines~\ref{ln:ex2DD:for} and \ref{ln:ex2DD:if}, we search in $Y^\text{AV}$ for the $y_i$ entities (\ie $\OOP{L}_i$ structures) involving $\mathbf{P}$ and another individual.
Line~\ref{ln:ex2DD:building} and \ref{ln:ex2DD:reading} perform the same operations done by the \CODE{AV.build} method but, in this case, it only involves the individual related to the identified entities $y_i$.
For example, if $Y^\text{AV}$ describes \ROLE{\alpha}{\beta}{\mathbf{P}}, \ROLE{\alpha}{\gamma}{\mathbf{P}} and \ROLE{\alpha}{\epsilon}{\mathbf{R}}, then the returned descriptors will be grounded on $\beta$ and $\gamma$.

Based on \OOP{H} and \OOP{B}, Example~\ref{alg:ex2} defines a function to make the robot aware of reachable locations given its IRI, \eg \CODE{I$_\rho$}~${=}$~\ONTO{`Robot1'}.
At Lines~\ref{ln:ex2:robotDescr1}--\ref{ln:ex2:robotDescr2}, the function creates an  individual descriptor \OOP{B} with $x=\IRI{\CODE{I$_\rho$}}$.
Then, \OOP{B} reads all axioms including the position of the robot represented through the property \ONTO{isIn}, which has only one entity since it is \emph{functional}.
At Line~\ref{ln:ex2:buildPos} such a position is used to build an individual descriptor having the ground representing the current robot location, \eg \ONTO{Corridor1}.

At Line~\ref{ln:ex2:connLoc}, we build the two individual Descriptors, which are related to \ONTO{Corridor1} through the \ONTO{isConnectedTo} property, \ie the locations \ONTO{Room1} and \ONTO{Room2}.
For each connected location, Line~\ref{ln:ex2:buildConn} builds descriptors with a ground equivalent to the classes whereby they are instances,  \ie $\{\top,$ \ONTO{LOCATION}, \ONTO{INDOOR}, $\ONTO{ROOM}\}$.
Line~\ref{ln:ex3:leaf} checks if those OWL classes are leaves in the TBox and, eventually, Line~\ref{ln:ex2:addPair} adds them to the output structure paired with the related individual, \eg  Line~\ref{ln:ex2:return} returns structures $\TUPLE{\ONTO{Room1},\ONTO{ROOM}}$ and $\TUPLE{\ONTO{Room2},\ONTO{ROOM}}$.

\input{./algorithm/example2.tex}

\subsection{Hierarchy of Compound Descriptors}
\label{sec:compaundExample}

In this use case, the robot has to continuously move among locations with open doors, check the status of the doors, and update the ontology accordingly.
Example~\ref{alg:ex3} presents a script implementing such operations in a scenario where the robot moves randomly across accessible locations.
In this case, we assume that the \ONTO{Robot1} can perceive whether a given \ONTO{DOOR} is open or closed (functionality defined in Line~\ref{ln:ex3:perceiveDef}), that it can cross doors (defined in Line~\ref{ln:ex3:move}), and that at least one door is open in each \ONTO{LOCATION}.

To address this use case, we used two OOP classes, \ie \OOP{W} and \OOP{V} defined in Example~\ref{alg:exExt3}, which exploit the compound individual descriptor \OOP{B} implemented in Example~\ref{alg:exDesF}.
The \OOP{W} descriptor extends \OOP{B} to encode axioms related to instances of the \ONTO{ROBOT} or \ONTO{DOOR} OWL classes, and it can differentiate the building method based on this difference.
Given \ROLE{\alpha}{\beta}{\mathbf{P}}, an instance of \OOP{W} with $x=\alpha$ builds \OOP{V} descriptors if \INST{\beta}{\ONTO{DOOR}} (as shown in Lines~\ref{ln:ex3:newBuild1}--\ref{ln:ex3:newBuild2}).
Otherwise, \OOP{W} builds \OOP{B} descriptors as specified in Line~\ref{ln:ex3D:buildSuper}.

\input{./algorithm/example3D-E.tex}

For showing purposes, we do not use the OWLOOP-based approach to determine whether $\beta$ is a \ONTO{DOOR}.
Instead, we use the factory-based paradigm provided by OWL-API, which is always supported by OWLOOP.
In particular, at Line~\ref{ln:ex3D:OWLfactory}, we query the ontology to classify the provided entities to build.
However, Example~\ref{alg:exExt3} simplifies such a classification since it would require constructing OWL axioms and querying the reasoner.

As shown in Line~\ref{ln:ex3D:TT}, \OOP{V} is a descriptor that \emph{extends} \OOP{W}, and \OOP{V} would always be constructed at Line~\ref{ln:ex3D:TTconstr} with a ground \INST{\beta}{\ONTO{DOOR}}.
For this reason, \OOP{V} can implement the function shown in Line~\ref{ln:ex3D:openClose}, which updates the state of a door.
In particular, Lines~\ref{ln:ex3D:open1}--\ref{ln:ex3D:open2} force the classification \INST{\beta}{OPEN} if the input parameter is \emph{true}; otherwise, we set \INST{\beta}{CLOSE}.
This function does not invoke the \CODE{write} method since the manipulations of the entity sets might require further processing before reasoning.

\input{./algorithm/example3.tex}

Example~\ref{alg:ex3} shows how \OOP{W} and \OOP{V} can be used to address the scenario presented at the beginning of this Section.
Lines~\ref{alg:ex3:const1}--\ref{alg:ex3:const2} get required entities from the ontology based on constant IRIs.
Line~\ref{ln:ex3:ssetup} constructs a full class descriptor \OOP{C} (defined in Section~\ref{sec:fullClassDescriptor}) with $x=\IRI{CLOSE}$, which is a new OWL class to be introduced in the ontology.
Line~\ref{ln:ex3:close2} creates the axiom $\ONTO{CLOSE}\sqsubseteq\ONTO{DOOR}$, and this is written in the ontology at Line~\ref{ln:ex3:close3}.
Then, Line~\ref{ln:ex3:open1} changes the ground of \OOP{C} to \ONTO{OPEN}, while its CS-based entity set still contains the axiom added at Line~\ref{ln:ex3:close2}.
Line~\ref{ln:ex3:open2} creates an OWLOOP axiom to specify that open and closed doors are disjoint classes, and the changes are written%
    \footnote{When a class $\Delta$ is asserted to be disjointed from another class $\Lambda$, the OWL-API automatically adds in the ontology two axioms for maintaining the consistency, \ie $\Delta=\neg\Lambda$ and $\Lambda=\neg\Delta$.} 
at Line~\ref{ln:ex3:esetup}.
Since we explicitly write all axioms, we do not reason and read inferred axioms.

Line~\ref{ln:ex3:robot} instantiates a descriptor \OOP{W} concerning \ONTO{Robot1}.
Within a loop, Line~\ref{ln:ex3:robotRead} reads axioms representing the robot, Line~\ref{ln:ex3:robotLocation} retrieves its current location, and Line~\ref{ln:ex3:visibleDoor} builds descriptors grounded on the related doors, \ie it builds instances of \OOP{V}, in accord with Example~\ref{alg:exExt3}.
Then, we iterate for each door that the robot can observe from its location.
At Line~\ref{ln:ex3:perceive}, the robot perceives doors states, which are updated in the ontology at Line~\ref{ln:ex3:updateStateWrite}.
Line~\ref{ln:ex3:openDoors} populates a list of open doors that the robot can cross and, at Lines~\ref{ln:ex3:smove}--\ref{ln:ex3:emove}, we choose a random door through which it will move.
When the robot reaches the new location, the process is repeated from Line~\ref{ln:ex3:loop}.

\section{Discussion}
\label{sec:discussion}

At the end of Section~\ref{sec:related_work}, we listed the features of OWLOOP (\ie $I, II, \ldots, VII$) that have been presented throughout the paper, and a discussion of our contribution follows.

\begin{enumerate}
    \item[$I$.]
Examples~\ref{alg:ex1}, \ref{alg:ex2} and \ref{alg:ex3} show how OWLLOP can \emph{read} axioms from an ontology.
In other words, OWLOOP encodes OWL entities as the attributes of an OOP object that is characterised by an OWL expression, \ie an object that implements the \emph{descriptor} interface detailed in Section~\ref{sec:OWLOOPDescriptor}.
    \item[$II$.]
Examples~\ref{alg:ex1} and \ref{alg:ex3} show how OWLOOP can \emph{write} OWL axioms in the ontology based on the entities encoded in a descriptor.
The reading and writing processes exploit a bijective map defined in Section~\ref{sec:mapping} for different OWL expressions.
    \item[$III$.]
Through $I$ and $II$, OWLOOP allows mapping the attributes of OOP objects with the entities of an OWL ontology at runtime. 
The formers are exploited by software components, while the latter are subjected to OWL reasoning.
As shown in Example~\ref{alg:ex1}, the reading, writing and reasoning processes are designed to be synchronised among OOP objects that concurrently modify the ontology and require reasoning.
    \item[$IV$.]
Examples~\ref{alg:exDes2} and \ref{alg:exDesF} show that the amount of OWL axioms mapped on OOP counterparts is customizable based on the application.
As detailed in Section~\ref{sec:implementation}, OWLOOP provides three types of descriptors interfaces, which map axioms based on modular combinations of OWL expressions.
This approach allows specifying the semantics of the axioms that a descriptor maps, and this might limit the computation load.
    \item[$V$.]
Examples~\ref{alg:ex2} and \ref{alg:ex3} show how an OWLOOP descriptor can \emph{build} other descriptors.
The built descriptors encode OWL entities related to the building descriptor through OWL axioms in the ontology.
This mechanism hides the \emph{factory} design paradigm of OWL-API since it allows \emph{getting} (and \emph{setting}) the axioms representing an OWL entity given an OOP object describing another OWL entity related to the former through an OWL expression.
Thus, OWLOOP allows the implementation of OOP objects mutually related based on the ontology.
This mechanism reduces the amount of constant IRIs that OOP objects require to access the knowledge in the ontology.
In addition, the OOP-based reading, writing, and building mechanisms of OWLOOP limit the paradigm shift required to use the OWL formalism since it avoids explicitly constructing OWL axioms.
    \item[$VI$.]
Example~\ref{alg:exExt3} shows that OWLOOP descriptors can be arranged within an OOP hierarchy of objects. 
Hence, it is possible to implement an OOP object that \emph{inherits} the ability to describe some OWL axioms from another object, which functionalities might be extended.
In other words, OWLOOP users can implement modular OOP architectures that exploit semantic representation and OWL reasoning.
    \item[$VII$.]
Since OWLOOP is based on OWL-API, which remains accessible to OWLOOP users (as shown in Examples~\ref{alg:ex1} and \ref{alg:exExt3}), any implementation based on OWL-API is compatible with systems exploiting OWLOOP.
This assures the compatibility of OWLOOP with all OWL reasoners, which expressiveness is not limited.
\end{enumerate}

As surveyed in Section~\ref{sec:related_work}, passive OWL to OOP mapping strategies (\eg the OWL-API) entirely support OWL reasoners, but they introduce a paradigm shift.
In contrast, active OWL to OOP mapping strategies limit the paradigm shift, but they affect reasoning capabilities.
To do not limit reasoning, OWLOOP implements a passive OWL to OOP mapping strategy based on OWL-API, and it uses the factory design pattern to exploit ontologies and reasoners.
Nonetheless, OWLOOP also maps OWL axioms through the descriptor interface, which manages the attributes of OOP objects used to represent some of the entities in the ontology.
Differently from OWL-API, these OOP objects support polymorphism, and they might be mutually related based on the ontology.
Thus, OWLOOP limits the paradigm shift since it provides OOP-like methods to manipulate and query OWL axioms subjected to reasoning, similarly to an active mapping strategy.

OWLOOP provides an OOP-like interface to the ontology based on the reading and writing methods, from which we derived the building method.
The latter is used to relate OOP objects representing OWL entities that are related in the ontology.
However, as introduced in Section~\ref{sec:axiomBuilding}, this mechanism has some limitations.
In particular, if a descriptor concerns OWL expressions that encode structures of OWL entities (\eg \OOP{L}), then the building methods will have multiple definitions.
Moreover, when a descriptor encodes a structure not known \emph{a priori} (\eg the most general representation of \OOP{R}), the implementation of the \CODE{build} method remains an open issue.
However, if we consider less general scenarios, \ie where some parts of the ontology are known \emph{a priori}, then the limitations of the building mechanism can be overcome.

Typical OWLOOP users are experts in OWL formalism and system developers.
OWL experts should represent prior knowledge in the ontology and develop appropriate descriptors based on the templates that OWLOOP provides, which have been presented in Section~\ref{sec:implementation}.
Instead, system developers should integrate ontologies and the related descriptors in a software architecture that exploits semantic knowledge representation and reasoning.
OWLOOP provides OWL experts with modular OOP objects for deploying ontologies while allowing system developers to avoid the paradigm shift.
Although the descriptors presented in Section~\ref{sec:implementation} could be used in an application, OWLOOP is designed to support developers when designing descriptors customised for their ontologies.
Customised descriptors are trivial to implement, and they can reduce the computation load when focusing on required OWL axioms only. 
Also, they can improve the readability of the code because the semantics of the encoded entities would be less abstract than general-purpose implementations, \eg the readability of Example~\ref{alg:exExt3} might be further improved if we consider less general descriptors.

This paper presents OWLOOP without discussing some OWL axioms, \ie class complement, class a value, disjoint union, negative assertion property, and chain property.
However, all those axioms can be addressed with further OWLOOP descriptors structured similarly to the ones discussed in this paper.
\cite{OWLOOP21} present a prototype implementation\textsuperscript{\ref{nt:github}} of OWLOOP that we used to validate our design for a subset of OWL axioms. 
Differently, this paper details the OWL to OOP mapping strategy of OWLOOP for being formally aligned with the UML design of OWL, presented by \cite{b2012OWL}.
Also, this paper specifies the OWLOOP implementation, which might also be based on a passive OWL to OOP mapping strategy different from OWL-API, \eg exploiting the Jena API proposed by \cite{JenaAPI}.
The prototyping implementation of OWLOOP simply considers OWL classes as defined from a conjunction of restrictions without a specific order.
In contrast, this paper presents the implementation of the OWLOOP restrictions \OOP{R}, which can be used to define OWL classes through conjunctions and disjunctions of restrictions with operation precedences.

OWLOOP, through its prototyping implementation, has been successfully used to deploy a framework for human activity recognition in smart homes and for developing an approach that learns classes in the TBox given knowledge over time,  as discussed by \cite{s2018Arianna} and \cite{oldSIT}, respectively. 
It has also been used in a human-robot collaboration scenario and for implementing a cognitive-like robot memory to support teaching by demonstration scenarios, as presented by \cite{l2018DialogueBased} and \cite{memoryLike}, respectively.
For those scenarios, we injected\footnote{The implementation is available at \url{https://github.com/buoncubi/ARMOR_OWLOOP}.} OWLOOP in a ROS Multi Ontology Reference (aRMOR), presented by \cite{armor_ws}, which provides services for OWL ontologies in software architectures based on the Robot Operating System (ROS).

\section{Conclusions}
\label{sec:conclusions}  

The paper presents the OWLOOP API that extends the OWL-API for using ontologies through OOP paradigms. 
In particular, OWLOOP allows getting and setting knowledge in the ontology, as it would be done for an OOP object, and exploits polymorphism to extend the knowledge representation in the OOP domain without limiting the reasoners.

OWLOOP performs a passive OWL to OOP mapping, and it provides novel interfaces to support the design and deployment of OWL ontology in OOP-based software architectures, especially by improving their modularity and reducing the paradigm shift.
Through examples based on a scenario where a robot moves in a smart environment, the paper shows the benefits and limitations of OWLOOP.

As future work, we aim at improving our prototype implementation\textsuperscript{\ref{nt:github}} to be fully compliant with the formalisation addressed in this paper. 
Furthermore, we aim at comparing the computational performance of OWL-API and OWLOOP, to quantify the complexity introduced by OWLOOP. 
Finally, we want to deploy OWLOOP in different scenarios to characterise development patterns when our novel API is adopted.

\bibliographystyle{./template/tlplike}
\bibliography{OWLOOP} 
\end{document}

%% file: numenclature.tex
\newcommand{\ONTO}[1]{\textsf{#1}\xspace}
\newcommand{\IRIdef}{\OWL{N}}
\newcommand{\IRI}[1]{\ensuremath{{\IRIdef(\ONTO{#1}})\xspace}}
\newcommand{\IRIentity}[2]{\ensuremath{{#1=\IRI{#2}}}}
\newcommand{\OWL}[1]{\ensuremath{{\mathscr{#1}}}\xspace}
\newcommand{\Expr}[1]{\ensuremath{{\OWL{E}_{#1}}}\xspace}
\newcommand{\ROLE}[3]{\ensuremath{{#1{,}#2{:}#3}}\xspace}%
\newcommand{\INST}[2]{\ensuremath{{#1{:}#2}}\xspace}%
\newcommand{\DOMAIN}[2]{\ensuremath{{#1\!\textbf{.}#2}}\xspace}%
\newcommand{\CODE}[1]{\texttt{#1}\xspace}
\newcommand{\OOP}[1]{\ensuremath{{\mathcal{#1}}}\xspace}
\newcommand{\IMPL}[2]{\ensuremath{{\OOP{#1}^{\OOP{#2}}}}\xspace}
\newcommand{\IMPLd}[1]{\IMPL{#1}{D}}
\newcommand{\OOPtempl}[2]{\ensuremath{{\OOP{#1}(#2)}}\xspace}
\newcommand{\TUPLE}[1]{\ensuremath{{\langle#1\rangle}}\xspace}
\newcommand{\OWLax}[2]{\TUPLE{\text{#1},\{#2\}}}
\newcommand{\tabHEADER}[2]{\multicolumn{#1}{c}{\emph{#2}}\\}

\newcommand{\cellNLl}[1]{\begin{oldtabular}[*]{@{}l@{}}#1\end{oldtabular}}
\newcommand{\cellTUPLE}[3]{\TUPLE{\text{#1},~#2,~#3}}
\newcommand{\THIS}[1]{\textnormal{\textbf{this}}\CODE{.}\!\!\!#1}%
\newcommand{\SUPER}[2]{\textnormal{\textbf{super}{}}\CODE{$($#2$)$}}%
\newcommand{\eg}{\emph{e}.\emph{g}.,~} 
\newcommand{\ie}{\emph{i}.\emph{e}.,~}

%% file: figure/figures_setup.tex
\newboolean{shouldegeneratefigure}
\setboolean{shouldegeneratefigure}{true}
\newcommand{\genericfigure}[2]{
    \ifthenelse{\boolean{shouldegeneratefigure}}{%
        \centering%
        \footnotesize%
        \def\svgwidth{#1\textwidth}%
        \setstretch{.95}%
        \input{./figure/#2.eps_tex}%
    }{%
        \centering%
        \includegraphics{./figure/generated/#2.pdf}%
    }%
}%
\newcommand{\FIGabstract}{\genericfigure{0.8}{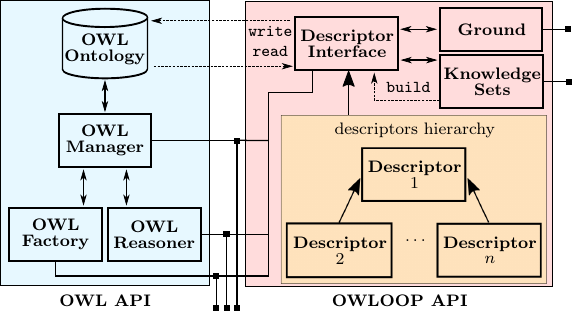}}%
\newcommand{\FIGontoExample}{\genericfigure{0.8}{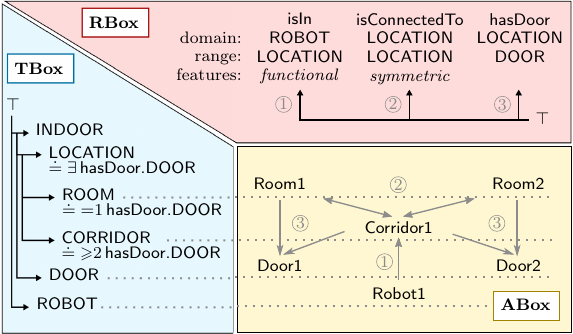}}%
\newcommand{\FIGowlentity}{\genericfigure{0.8}{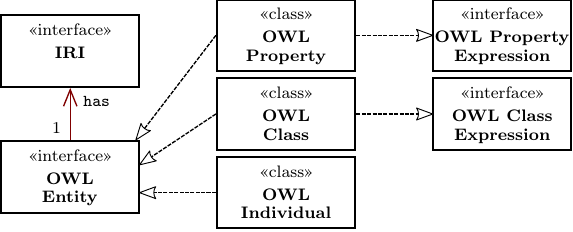}}%
\newcommand{\FIGowlproperty}{\genericfigure{0.8}{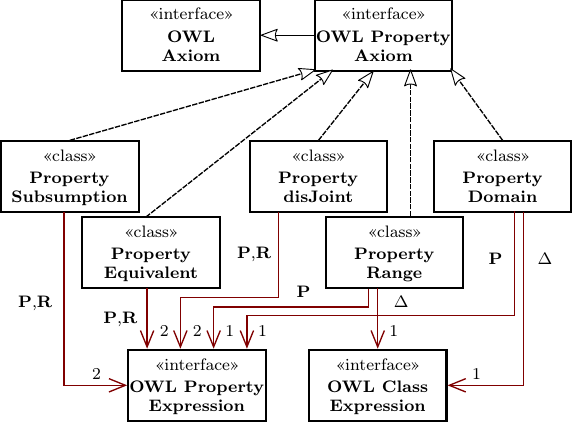}}%
\newcommand{\FIGowlpropertybis}{\genericfigure{0.8}{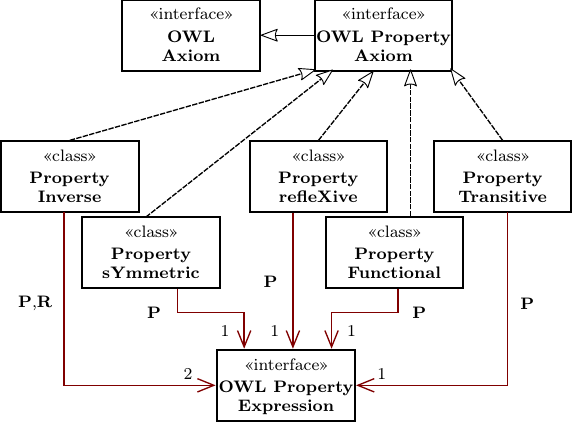}}%
\newcommand{\FIGowlclassaxiom}{\genericfigure{0.8}{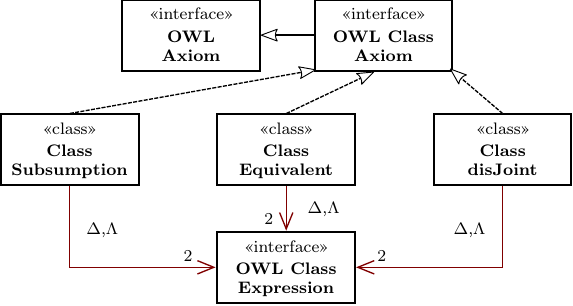}}%
\newcommand{\FIGowlclassexpression}{\genericfigure{0.8}{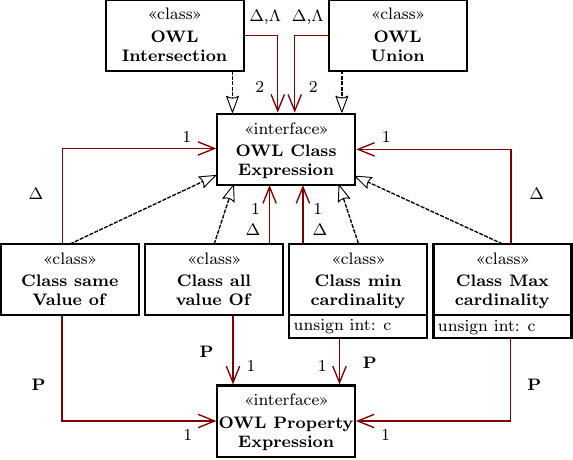}}%
\newcommand{\FIGowlassertion}{\genericfigure{0.8}{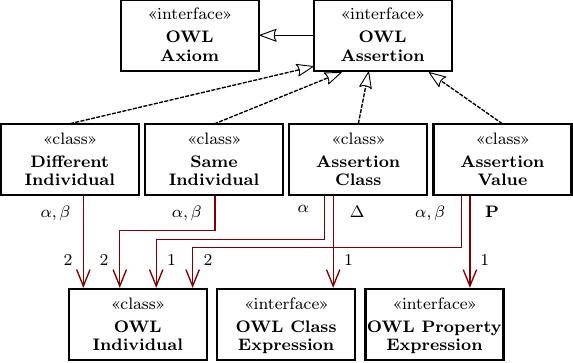}}%
\newcommand{\FIGowloopentity}{\genericfigure{0.8}{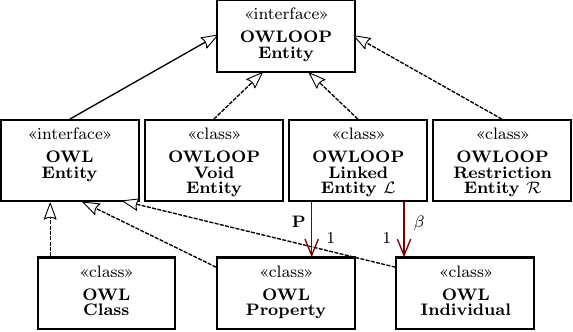}}%
\newcommand{\FIGowlooprestriction}{\genericfigure{0.8}{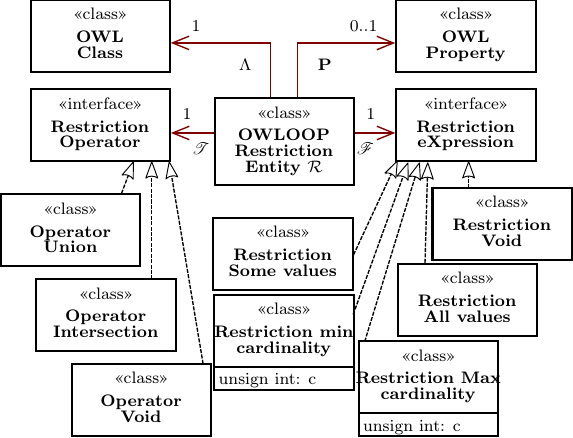}}%
\newcommand{\FIGdescriptorproperty}{\genericfigure{0.8}{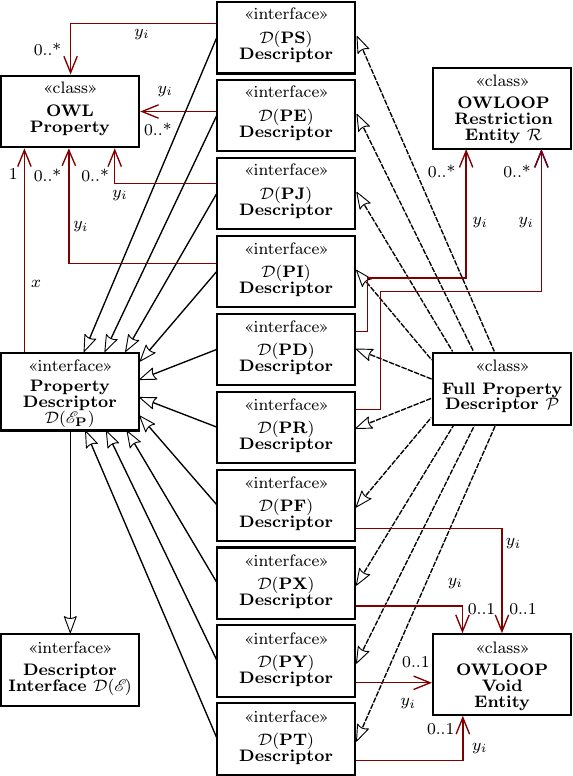}}%
\newcommand{\FIGdescriptorclass}{\genericfigure{0.8}{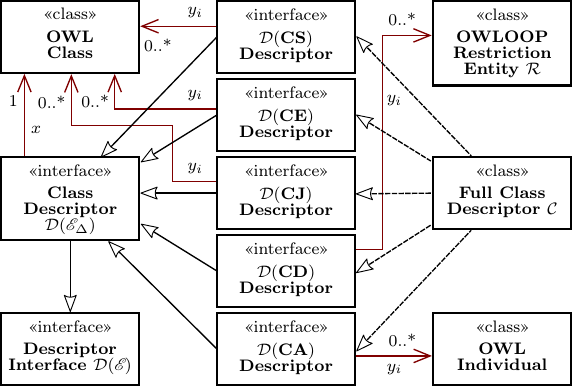}}%
\newcommand{\FIGdescriptorindividual}{\genericfigure{0.8}{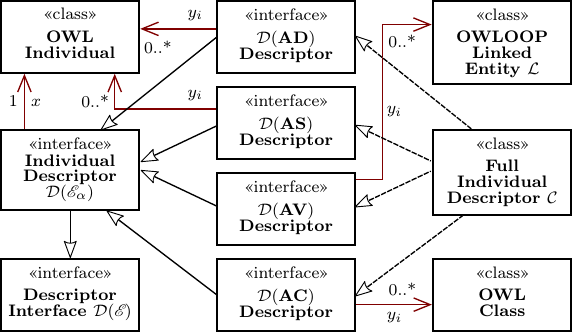}}%
\newcommand{\FIGsummaryclass}{\genericfigure{.09}{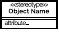}}%
\newcommand{\FIGsummaryattribute}{\genericfigure{0.07}{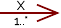}}%
\newcommand{\FIGsummaryextension}{\genericfigure{.07}{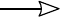}}%
\newcommand{\FIGsummaryrealization}{\genericfigure{.07}{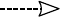}}%

%% file: table/acronym.tex
\footnotesize
\centering
{\tablefont\begin{tabular}{@{\extracolsep{\fill}}rlcrlcrlcrl}
\topline
\tabHEADER{11}{OWL Property}
    PS:\,  &  Subsumption,     &~~& PJ:\,  &  Equivalence,     &~~& PE:\,  &  disJunction,     &~~& PI:\,  &  Inverse,              \\
    PD:\,  &  Domain,          &  & PR:\,  &  Range,           &  & PF:\,  &  Functional,      &  & PX:\,  &  refleXive,            \\
    PY:\,  &  sYmmetric,       &  & PT:\,  &  Transitive.      &  & LE:\,  &  \multicolumn{3}{l}{OWLOOP Linked Entitiy.}            %
\midline        
\tabHEADER{11}{OWL Class}
    CE:\,  &  Equivalence,     &  & CJ:\,  &  disJunction,     &  & CS:\,  &  Subsumption,     &  & CD:\,  &  Declaration,          \\
    CI:\,  &  Intersection,    &  & CU:\,  &  Union,           &  & CV:\,  &  some Value of,   &  & CO:\,  &  all value Of,         \\
    Cm:\,  &  min cardinality, &  & CM:\,  &  Max cardinality. &  & CA:\,  &  \multicolumn{3}{l}{OWLOOP Class Assertion.}           %
\midline
\tabHEADER{11}{OWLOOP Restriction}
    RE:\,  &  Entity,          &  & RO:\,  &  Operator,        &  & RX:\,  &  eXpression,      &  & RV:\,  &  Void,                 \\
    RA:\,  &  All values of,   &  & RS:\,  &  Some value of,   &  & Rm:\,  &  min cardinality, &  & RM:\,  &  Max cardinality.      %
\midline
\tabHEADER{11}{OWLOOP Operator}
    OI:\,  &  Intersection,    &  & OU:\,  &  Union,           &  & OV:\,  &  Void.            &  &      &                          %
\midline
\tabHEADER{11}{OWL Assertion}
    AC:\,  &  Class,           &  & AV:\,  &  Value,           &  & AS:\,  &  Same individual, &  & AD:\,  &  Different individual. %
\botline
\end{tabular}}

%% file: table/glossary.tex
\footnotesize
\centering
{\tablefont\begin{tabular}{@{\extracolsep{\fill}}rl@{}c@{}rl}
\topline
\tabHEADER{5}{General Notation for Knowledge in the OWL Ontology}
    \emph{Bold Roman}                & OWL property (\eg $\mathbf{P},\mathbf{R}$).     & ~ & 
    \emph{Greek}                     & OWL classes (\eg $\Delta,\Lambda$).             \\
    \emph{greek}                     & OWL individual (\eg $\alpha,\beta$).            & &
    \ONTO{Font}                      & Identifies IRI.                                 \\        
    \IRI{IRI}                        & OWL entity specified by IRI.                    & &
    \ONTO{camelCase}                 & OWL Property IRI.                               \\
    \ONTO{UPPER\,CASE}               & OWL Class IRI.                                  & &
    \ONTO{Capitalised}               & OWL Individual IRI.                             \\
    \emph{mathscr}                   & \multicolumn{3}{l}{Font for semantic structures, 
                                        \eg \OWL{E}, \OWL{A}, etc.}                    \\%
\tabHEADER{5}{Symbols About Ontological Knowledge}
    \OWL{O}                          & OWL Ontology.                                   & &
    \OWL{E}                          & OWL Expression.                                 \\
    \Expr{\mathbf{P}}                & Property Expression.                            & &
    \Expr{\Delta}                    & Class Expression.                               \\
    \Expr{\alpha}                    & Assertion.                                      & &
    $\Pi$                            & Anonymous OWL Class.                            \\
    \OWL{A}                          & OWL Axiom.                                      & &
    \OWL{P}                          & OWLOOP Axiom.                                   \\
    \OWL{T}                          & OWLOOP Operator.                                & &
    \OWL{F}                          & OWLOOP Expression.                              %
\midline
\tabHEADER{5}{General OOP Notation}
    \CODE{font}                      & Identifies OOP elements.                        & &
    \CODE{obj.method$()$}            & OOP member specifier.                           \\
    \emph{mathcal}                   & \cellNLl{Font for OOP interfaces,\\
                                            \eg $\OOP{D}$, $\OOP{R}$, etc.}            & &         
    \IMPLd{T}                        & \cellNLl{Superscript identifies an\\ 
                                            object implementing $\OOP{D}$.}            \\
    \OOPtempl{T}{t}                  & Template specification.                         & &
    \TUPLE{a,b,\ldots}               & OOP members structure.                          \\%
\tabHEADER{5}{Symbols About OOP Structures}
    \OOP{O}                          & OWL Ontology API.                               & &
    $\emptyset$                      & A \CODE{void} structure.                        \\
    \OOP{D}                          & OWLOOP descriptor.                              & & 
    \OOP{R}                          & OWLOOP restriction.                             \\
    \OOP{L}                          & OWLOOP link.                                    & & 
    \OOP{I}                          & Mapping Intent.                                 \\
    $\OOP{D}(\Expr{\mathbf{P}})$     & Property descriptor.                            & & 
    $\OOP{D}(\Expr{\Delta})$         & Class descriptor.                               \\
    $\OOP{D}(\Expr{\alpha})$         & Individual descriptor.                          & & 
    \OOP{P}                          & Full Property descriptor.                       \\
    \OOP{C}                          & Full Class descriptor.                          & & 
    \OOP{A}                          & Full Individual descriptor.                     %
\midline
\tabHEADER{5}{General Set and Elements Notation}
    \emph{Roman}                     & \cellNLl{Sets, having coherent\\ 
                                            elements (\eg $e_i\in E$).}                & &
    \emph{roman}                     & \cellNLl{Set elements, numbers,\\ 
                                            or indexes (\eg $e,i$).}                   \\%
\tabHEADER{5}{Symbols About Sets and Elements}
    $E$                              & Entities of \OWL{A}.                            & &         
    $x$                              & Descriptor \OOP{D} ground.                      \\
    $y_i\in Y$                       & The entity set of \OOP{D}.                      & &
    $P$                              & Set of all $\OWL{E}_\mathbf{P}$ expressions.    \\
    $C$                              & Set of all $\OWL{E}_\Delta$ expressions.        & &       
    $A$                              & Set of all $\OWL{E}_\alpha$ expressions.        \\
    $Z$                              & \multicolumn{3}{l}{Members of $\OOP{R}$.}       %
\midline
\tabHEADER{5}{UML Conventions}
    \cellNLl{~\\\FIGsummaryclass}
                                     & \cellNLl{Class, attributse are shown\\
                                             only if necessary.}                       & &
    \cellNLl{~\\[-.5em]\FIGsummaryattribute}
                                     & \cellNLl{Attributes association,\\
                                             name and cardinality.}                    \\ 
    \FIGsummaryextension
                                     & Class extension.                                & &
    \FIGsummaryrealization
                                     & Interface realisation.                          %
\botline
\end{tabular}}

%% file: table/owlAxiom.tex
\footnotesize
\centering
{\tablefont\begin{tabular}{@{\extracolsep{\fill}}rlcrl}
\topline
\tabHEADER{5}{Property Expression ($\mathcal{E}_\mathbf{P}$) Axioms in the RBox}
    \texttt{Subsumption}          (PS)   & $\mathbf{P}\sqsubseteq\mathbf{R}$                          & ~~~ &
    \texttt{disJoint}             (PJ)   & $\mathbf{P}=\neg\mathbf{R}$                                \\
    \texttt{Equivalent}           (PE)   & $\mathbf{P}\equiv\mathbf{R}$                               & &
    \texttt{Inverse}              (PI)   & $\mathbf{P}=\mathbf{R^{-1}}$                               \\
    \texttt{Domain}               (PD)   & \DOMAIN{\mathbf{P}}{\Delta}                                & & 
    \texttt{Range}                (PR)   & \DOMAIN{\mathbf{P^{-1}}}{\Delta}                           \\%
\tabHEADER{5}{Property Features}
    \texttt{Functional}           (PF)   & $\exists\,\mathbf{P}\sqsubseteq{<}1\mathbf{P}$             & &
    \texttt{refleXive}            (PX)   & $\top\sqsubseteq\exists\,\DOMAIN{\mathbf{P}}{\ONTO{Self}}$ \\
    \texttt{sYmmetric}            (PY)   & $\mathbf{P}\sqsubseteq\mathbf{P^{-1}}$                     & &
    \texttt{Transitive}           (PT)   & $\mathbf{P^{+}}\sqsubseteq\mathbf{P}$                      %
\midline
\tabHEADER{5}{Class Expression ($\mathcal{E}_\Delta$) Axioms in the TBox}           
    \texttt{Equivalent}           (CE)   & $\Delta\equiv \Lambda$                                     & &
    \texttt{disJoint}             (CJ)   & $\Delta=\neg\Lambda$                                       \\
    \texttt{Subsumption}          (CS)   & $\Delta\sqsubseteq \Lambda$                                & &
    \texttt{Declaration}          (CD)   & $\Delta\doteq \Lambda$                                     \\%
\tabHEADER{5}{Class Operators}
    \texttt{Intersection}         (CI)   & $\Delta\sqcap\Lambda$                                      & &
    \texttt{Union}                (CU)   & $\Delta\sqcup\Lambda$                                      \\ 
    \texttt{some Values of}       (CV)   & $\exists\,\DOMAIN{\mathbf{P}}{\Delta}$                     & &
    \texttt{all values Of}        (CO)   & $\forall\,\DOMAIN{\mathbf{P}}{\Delta}$                     \\
    \texttt{min Cardinality}      (Cm)   & ${\geqslant}c\,\DOMAIN{\mathbf{P}}{\Delta}$                & &
    \texttt{Max Cardinality}      (CM)   & ${\leqslant}c\,\DOMAIN{\mathbf{P}}{\Delta}$                %
\midline
\tabHEADER{5}{Individual Assertion ($\mathcal{E}_\alpha$) Axioms in the ABOX}
    \texttt{Class}                (AC)   & \INST{\alpha}{\Delta}                                      & &
    \texttt{Value}                (AV)   & \ROLE{\alpha}{\beta}{\mathbf{P}}                           \\
    \texttt{Same individual}      (AS)   & $\alpha = \beta$                                           & &
    \texttt{Different individual} (AD)   & $\alpha \neq \beta$                                        %
\botline
\end{tabular}}

%% file: algorithm/Dinterface.tex
\DecMargin{\ALGOINDENT}
\begin{pseudocode}[!t]
   \caption{The descriptor interface.}
   \label{alg:OWLOOPdescrInterface}%
\footnotesize
    \SetKwInOut{Expr}{$\mathscr{E}$}\Expr{Extends \CODE{OWLExpression}.}
    \SetKwInOut{Gnd}{$x$}\Gnd{Extends \CODE{OWLEntity} (ground type, derived from \OWL{E}).}
    \SetKwInOut{Entity}{$y_i$}\Entity{Extends \CODE{OWLOOPEntity} (entity type, derived from \OWL{E}).}
    \SetKwInOut{EntitySet}{$Y$}\EntitySet{\CODE{Set} of $y_i$ elements without repetitions, with $i\in[1,n]$.}
    \SetKwBlock{interf}{\Numberline interface \textnormal{\OOPtempl{D}{\OWL{E}}}\hfill}{}
\vspace{.5em}
    \interf(\hfill\tcp*[h]{\footnotesize \OWL{E} defines $x$ and $y_i$ template parameters.}\label{ln:expression}){
        \Numberline\OOP{O} getOntology()\label{ln:getontology}\hfill\tcp{\footnotesize Immutable reference to an API that manages OWL ontologies.}
\vspace{.8em}
        \Numberline$Y$ getEntities()\label{ln:getEntities}\hfill\tcp{\footnotesize Immutable reference to the entity set.}
\vspace{.8em}
        \Numberline$x$ getGround()\label{ln:getGround}\hfill\tcp{\footnotesize Ground getter.}
        \Numberline\textbf{void} setGround($x$ ground)\label{ln:setGround}\hfill\tcp{\footnotesize Ground setter.}
\vspace{.8em}
        \Numberline\textbf{protected} \emph{OWLAxiom}$[\,]$ query()\label{ln:query}$\{\ldots\}$\hfill\tcp{\footnotesize Based on \ref{ln:getontology}. Used by \ref{ln:read} and \ref{ln:write}.}
\vspace{.8em}  
        \Numberline $\OOP{I}[\,]$ read()\label{ln:read}$\{\ldots\}$\hfill\tcp{\footnotesize OWL to OWLOOP axioms mapping (changes $Y$).}
        \Numberline $\OOP{I}[\,]$ write()\label{ln:write}$\{\ldots\}$\hfill\tcp{\footnotesize OWLOOP to OWL axioms mapping (changes \OWL{O}).}
\vspace{.8em}
        \Numberline \textbf{protected} \OOP{D} newEntityToBuild($y_i$ entity)\;\label{ln:newbuild}
        \Numberline \OOP{D[\,]} build()\label{ln:build}$\{\ldots\}$\hfill\tcp{\footnotesize It invokes \ref{ln:newbuild} and \ref{ln:read} for each entity in \ref{ln:getEntities}.}
}
\end{pseudocode}
\IncMargin{\ALGOINDENT}

%% file: table/owloopAxiom.tex
\footnotesize
\centering
{\tablefont\begin{tabular}{@{\extracolsep{\fill}}rlcrlcrl}
\topline
 \tabHEADER{8}{OWLOOP Property Axiom Description ($\mathcal{E}_\mathbf{P}$)}
    \texttt{Subsumption}:          & \cellTUPLE{PS}{\mathbf{P}}{\mathbf{R}}           & ~~ & 
    \texttt{Equivalent}:           & \cellTUPLE{PE}{\mathbf{P}}{\mathbf{R}}           & ~~ &
    \texttt{disJoint}:             & \cellTUPLE{PJ}{~\!\mathbf{P}}{\mathbf{R}}        \\ 
    \texttt{Inverse}:              & \cellTUPLE{PI}{~\!\mathbf{P}}{\mathbf{R^{-1}}}   &&
    \texttt{Range}:                & \cellTUPLE{PR}{\mathbf{P}}{\OOP{R}}              &&
    \texttt{Domain}:               & \cellTUPLE{PD}{\mathbf{P}}{\OOP{R}}              \\
    \texttt{Functional}:           & \cellTUPLE{PF}{\mathbf{P}}{\emptyset}            && 
    \texttt{refleXive}:            & \cellTUPLE{PX}{\mathbf{P}}{\emptyset}            &&
    \texttt{sYmmetric}:            & \cellTUPLE{PY}{\mathbf{P}}{\emptyset}            \\
    \texttt{Transitive}:           & \cellTUPLE{PT}{\mathbf{P}}{\emptyset}            %
\midline
\tabHEADER{8}{OWLOOP Class Axiom Description ($\mathcal{E}_\Delta$)}
    \texttt{Declaration}:          & \cellTUPLE{CD}{\Delta}{\OOP{R}}                 && 
    \texttt{Subsumption}:          & \cellTUPLE{CS}{\Delta}{\Lambda}                 &&
    \texttt{Equivalent}:           & \cellTUPLE{CE}{\Delta}{\Lambda}                 \\ 
    \texttt{disJoint}:             & \cellTUPLE{CJ}{~\Delta}{\Lambda}                &&
    \texttt{Assertion}:            & \cellTUPLE{CA}{\!\Delta}{\alpha}                  
\midline
\tabHEADER{8}{OWLOOP Individual Assertion Description ($\mathcal{E}_\alpha$)}       
    \texttt{Class}:                & \cellTUPLE{AC}{\alpha}{\Delta}                  &&
    \texttt{Same individual}:      & \cellTUPLE{AS}{\alpha}{\beta}                   \\
    \texttt{Value}:                & \cellTUPLE{AV}{\alpha}{\OOP{L}}                 &&        
    \texttt{Different individual}: & \cellTUPLE{AD}{\!\alpha}{\beta}                 %
\botline
\end{tabular}}

%% file: algorithm/example1.tex
\newcommand\CODEsep{\vspace{.5em}}%
\begin{examplecode}[!t]%
    \caption{
    A function (based on the ontology shown in Figure~\ref{fig:owloop:example}) that creates a new location $\alpha$ having a door $\gamma$ connected to a known location $\beta$.}%
    \label{alg:ex1}%
    \footnotesize%
        \KwIn{The ontology interface \OOP{O}, and the IRIs \CODE{I$_\alpha$}, \CODE{I$_\beta$}, \CODE{I$_\gamma$} as textual strings.}%
        \SetKwInOut{KwConstant}{Constant}\KwConstant{A string representing the IRI \CODE{I$_\mathbf{P}$}~${=}$~\ONTO{`hasDoor'}.}%
        \KwOut{The OWL classes having $\alpha$ as an instance.}%
    \CODEsep%
        \tcp{Get OWL entities given their IRI (\ie through \IRIdef applied to the ontology).}%
        \CODE{OWLIndividual} $\alpha     \gets$ \CODE{\OOP{O}.individual$($I$_\alpha)$}\label{ln:ex1:IRIground1}\;
        \CODE{OWLIndividual} $\beta      \gets$ \CODE{\OOP{O}.individual$($I$_\beta)$}\;
        \CODE{OWLIndividual} $\gamma     \gets$ \CODE{\OOP{O}.individual$($I$_\gamma)$}\;%
        \CODE{OWLProperty} $\mathbf{P} \gets$ \CODE{\OOP{O}.property$($I$_\mathbf{P})$}\label{ln:ex1:IRIground2}\;
    \CODEsep%
        \tcp{Create an OWLOOP individual descriptor to represent the new location $\alpha$.}%
        \OOP{A} \CODE{locationDescr} $\gets$ \textbf{new} \OOP{A(\OOP{O})}\;\label{ln:ex1:LocDescr1}%
        \CODE{locationDescr.setGround$(\alpha)$}\;\label{ln:ex1:LocDescr2}%
    \CODEsep%
        \tcp{Create an OWLOOP individual descriptor representing the location $\beta$.}%
        \OOP{A} \CODE{connectedDescr} $\gets$ \textbf{new} \OOP{A(\OOP{O})}\;\label{ln:ex1:ConDescr1}%
        \CODE{connectedDescr.setGround$(\beta)$}\;\label{ln:ex1:ConDescr2}%
    \CODEsep%
        \tcp{Add LE-based OWLOOP axioms to represent that $\alpha$ and $\beta$ locations share a door.}%
        \CODE{locationDescr.AV.getEntities$()$.add$(\mathbf{P},\gamma)$}\;\label{ln:ex1:Prop1}%
        \CODE{connectedDescr.AV.getEntities$()$.add$(\mathbf{P},\gamma)$}\;\label{ln:ex1:Prop2}%
    \CODEsep%
        \tcp{Map OWLOOP axioms into OWL axioms that are added to the ontology.}%
        \CODE{locationDescr.write$()$}\;\label{ln:ex1:Locwrite}%
        \CODE{connectedDescr.write$()$}\;\label{ln:ex1:Conwrite}%
    \CODEsep%
        \tcp{Perform OWL reasoning and query inferences.}%
        \CODE{\OOP{O}.reason$()$}\;\label{ln:ex1:reason}
        \CODE{locationDescr.read$()$}\;\label{ln:ex1:read}%
        \Return{\textnormal{\CODE{locationDescr.AC.getEntities$()$}}} \hfill \tcp{e.g., \ONTO{\{$\top$, LOCATION, INDOOR, ROOM\}}.\label{ln:ec1:out}}%
\end{examplecode}%

%% file: algorithm/example2D-H.tex
\begin{examplecode}[!t]%
    \caption{The implementation of an OWLOOP class descriptor that represents subsuming OWL axioms and identifies if the ground is a leaf node in the TBox.}%
    \label{alg:exDes2}%
    \footnotesize%
        \SetKwInOut{KwAttribute}{Attributes}\KwAttribute{The ontology interface \OOP{O}, a ground OWL Class $x$, and an empty entity set  $Y^{\text{CS}}$ that will contain OWL classes subsuming $x$.}%
    \CODEsep%
        \SetKwBlock{Constr}{constructor \OOP{H}\!\!(\CODE{Ontology ontology, OWLClass ground})}{}%
        \SetKwBlock{GetOntology}{override \textnormal{\CODE{Ontology} getOntology}$()$}{}%
        \SetKwBlock{GetGround}{override \textnormal{\CODE{OWLClass} getGround}$()$}{}%
        \SetKwBlock{SetGround}{override \textnormal{\CODE{void} setGround\CODE{$($OWLClass ground$)$}}}{}%
%
        \SetKwBlock{CSGetEntities}{override \textnormal{\CODE{OWLClass$[\,]$} CS.getEntities}$()$}{}%
        \SetKwBlock{CSSetEntities}{override \textnormal{\CODE{void} CS.setEntities$($\CODE{OWLClass$[\,]$} $Y)$}}{}%
        \SetKwBlock{CSBuild}{override \textnormal{\OOP{H} CS.newEntityToBuild}$($\CODE{OWLCLass} $y_i)$}{}%
%
        \SetKwBlock{IsLeaf}{\textnormal{\CODE{boolean} isLeaf}$()$}{}%
        \SetKwBlock{Interf}{class \textnormal{\OOP{H}} implements \textnormal{$\OOP{D}(\text{CS})$}}{}%

       \Interf(\label{ln:ex2:implements}){
            \Constr(\label{ln:ex2:constr}){%
                \THIS{\OOP{O}} $\gets$ \CODE{ontology},~~~~~
                \THIS{$x$} $\gets$ \CODE{ground},~~~~~
                \THIS{$Y^{\text{CS}}$} $\gets [\,]$\label{ln:ex2:entityInit}\; 
            }%
            \GetOntology{\label{ln:ex2:getO}%
                \Return{\THIS{\OOP{O}}}\;
            }%
            \GetGround{%
                \Return{\THIS{$x$}}\;\label{ln:ex2:getX}%
            }%
            \SetGround{%
                \THIS{$x$} $\gets$ \CODE{ground}\;\label{ln:ex2:setX}%
            }%
            \CSGetEntities{%
                \Return{\THIS{$Y^{\text{CS}}$}}\;\label{ln:ex2:getY}%
            }%
            \CSBuild{\label{ln:ex2:newBuild}%
                \Return{\textnormal{\CODE{\textbf{new} \OOP{H}\!\!$(\OOP{O},y_i)$}}}
            }%
        \CODEsep%
            \IsLeaf{\label{ln:ex2:leaf}%
                \Return{\textnormal{\THIS{CS.getEntities().length}$() = 1$ ~~and~~ \THIS{CS.getEntities}$()[0] = \bot$}}\;
            }%
        }%
\end{examplecode}%

%% file: algorithm/example2D-F.tex
\begin{examplecode}[!t]%
    \caption{The implementation of an OWLOOP compound individual descriptor representing OWL assertions involving object properties and classes.}%
    \label{alg:exDesF}%
    \footnotesize%
        \SetKwInOut{KwAttribute}{Attributes}\KwAttribute{The ontology interface $\OOP{O}$, an OWL Individual $x$, and two empty entity sets containing OWL Classes ($Y^{\text{AC}}$) and OWLOOP Links ($Y^{\text{AV}}$).}
    \CODEsep%
        \SetKwBlock{Constr}{constructor \OOP{B}\!\!(\CODE{Ontology ontology, OWLIndividual ground})}{}%
        \SetKwBlock{GetOntology}{override \textnormal{\CODE{Ontology} getOntology}$()$}{}%
        \SetKwBlock{GetGround}{override \textnormal{\CODE{OWLIndividual} getGround}$()$}{}%
        \SetKwBlock{SetGround}{override \textnormal{\CODE{void} setGround\CODE{$($OWLIndividual ground$)$}}}{}%
%
        \SetKwBlock{AVGetEntities}{override \textnormal{\CODE{$\OOP{L}[\,]$} AV.getEntities}$()$}{}%
        \SetKwBlock{AVSetEntities}{override \textnormal{\CODE{void} AV.setEntities$($\CODE{$\OOP{L}[\,]$} $Y)$}}{}%
        \SetKwBlock{AVBuildRange}{override \textnormal{\OOP{B} AV.newEntityToBuild}$($\CODE{OWLIndividual} $y_i)$\label{ln:ex2DD:newB}}{}%
        \SetKwBlock{AVBuildProperty}{override \textnormal{\OOP{P} AV.newPropertyToBuild}$($\CODE{OWLProperty} $y_i)$\label{ln:ex2DD:newP}}{}%
        \SetKwBlock{ACGetEntities}{override \textnormal{\CODE{OWLClass$[\,]$} AC.getEntities}$()$}{}%
        \SetKwBlock{ACSetEntities}{override \textnormal{\CODE{void} AC.setEntities$($\CODE{OWLClass$[\,]$} $Y)$}}{}%
        \SetKwBlock{ACBuild}{override \textnormal{\OOP{H} AC.newEntityToBuild}$($\CODE{OWLClass} $y_i)$\label{ln:ex2DD:newH}}{}%
        \SetKwBlock{Read}{override \textnormal{\OOP{I}\!\!$[\,]$ \CODE{read$()$}}}{}%
        \SetKwBlock{Write}{override \textnormal{\OOP{I}\!\!$[\,]$ \CODE{write$()$}}}{}%
        \SetKwBlock{Query}{override \textnormal{\CODE{OWLAxiom}\!\!$[\,]$ \CODE{query$()$}}}{}%
        \SetKwBlock{BuildRange}{\textnormal{\OOP{B}\!\!$[\,]$ AV.build}\CODE{$($OWLProperty property$)$}}{}%
%
        \SetKwBlock{Interf}{class \textnormal{\OOP{B}} implements \textnormal{$\OOP{D}(\text{AV})$, $\OOP{D}(\text{AC})$}}{}%
       \Interf(\label{ln:ex3:start}){%
            \Constr{%
                \THIS{\OOP{O}} $\gets$ \CODE{ontology},~~~~~
                \THIS{$x$} $\gets$ \CODE{ground},~~~~~
                \THIS{$Y^{\text{AC}}$} $\gets [\,]$,~~~~~
                \THIS{$Y^{\text{AV}}$} $\gets [\,]$\;
            }%
            \GetOntology{%
                \Return{\THIS{\OOP{O}}}\;
            }%
            \GetGround{%
                \Return{\THIS{$x$}}\;
            }%
            \SetGround{%
                \THIS{$x$} $\gets$ \CODE{ground}\;
            }%
\CODEsep%
            \AVGetEntities{%
                \Return{\THIS{$Y^{\text{AV}}$}}\;
            }%
            \ACGetEntities{%
                \Return{\THIS{$Y^{\text{AC}}$}}\;%
            }%
\CODEsep%
            \AVBuildRange{%
                \Return{\textnormal{\textbf{new} \OOP{B}$(\OOP{O}, y_i)$}}\label{ln:ex2:newBuild1}\;%
            }%
            \AVBuildProperty{%
                \Return{\textnormal{\textbf{new} \OOP{P}$(\OOP{O}, y_i)$}}\label{ln:ex2:newBuild2}\;%
            }%
 
            \ACBuild{
                \Return{\textnormal{\textbf{new} \OOP{H}$(\OOP{O},y_i)$}}\label{ln:ex3:end}\;%
            }%
\CODEsep%
            \Read(\label{ln:ex3:combine1}){%
                \CODE{\OOP{I}\!\!$[\,]$ intents $\gets [\,]$}\;%
                \CODE{intents.addAll}{(\THIS{\CODE{AC.read$()$}})},~~~~~
                \CODE{intents.addAll}{(\THIS{\CODE{AV.read$()$}})}\;%
                \Return{\textnormal{\CODE{intents}}}%
            }%
            \Write{%
                \CODE{\OOP{I}\!\!$[\,]$ intents $\gets [\,]$}\;%
                \CODE{intents.addAll}{(\THIS{\CODE{AC.write$()$}})},~~~~~
                \CODE{intents.addAll}{(\THIS{\CODE{AV.write$()$}})}\;%
                \Return{\textnormal{\CODE{intents}}}%
            }%
            \Query{%
                \CODE{OWLAxiom$[\,]$ axioms $\gets [\,]$}\;%
                \CODE{axioms.addAll}{(\THIS{\CODE{AV.query$()$}})},~~~~~
                \CODE{axioms.addAll}{(\THIS{\CODE{AC.query$()$}})}\;%
                \Return{\textnormal{\CODE{axioms}}}\label{ln:ex3:combine2}%
            }%
\CODEsep
            \BuildRange(\label{ln:ex2DD:buildRangeProp}){%
                \CODE{\OOP{B}\!\!$[\,]$ out $\gets$ $[\,]$}\;
                \ForEach{\label{ln:ex2DD:for}\textnormal{\CODE{\OOP{L} $y_i \in $ \THIS{AV.getEntities()}}}}{%
                    \If{\label{ln:ex2DD:if}\textnormal{\CODE{$y_i.\text{getProperty}() = \text{property}$}}}{%
                        \tcp{Instanciate and read (\ie \CODE{build}) \OOP{B} for a given $y_i$.}%
                        \CODE{built} $\gets$ \THIS{\CODE{AV.newEntityToBuild($y_i.\text{getIndividual}()$)}}\label{ln:ex2DD:building}\;%
                        \CODE{built.read$()$}\label{ln:ex2DD:reading}\;%
                        \CODE{out.add$($built$)$}\label{ln:ex2DD:storing}\;%
                    }%
                }%
                \Return \textnormal{\CODE{out}}\label{ln:ex2DD:ePropBuild}%
            }%
            
        }%
\end{examplecode}%

%% file: algorithm/example2.tex
\begin{examplecode}[!t]
    \caption{A function (based on the ontology shown in Figure~\ref{fig:owloop:example}) to find reachable locations.}%
    \label{alg:ex2}%
    \footnotesize
        \KwIn{The IRI idetifying a robot \CODE{I$_\rho$}.}
        \KwOut{Pairs of OWL entities related to classes and their realizations (\ie individuals) describing the location that the robot can reach.}    
    \CODEsep%
        \tcp{Acquire knowledge about the robot position.}%
        \CODE{OWLIndividual} $\rho \gets \CODE{\OOP{O}.individual$($I$_\rho)$}$\;
        \OOP{B} \CODE{robotDescr} $\gets$ \textbf{new} \OOP{B}\!\!$(\OOP{O}, \rho)$\;\label{ln:ex2:robotDescr1}%
        \CODE{robotDescr.read$()$}\;\label{ln:ex2:robotDescr2}%
        \CODE{OWLProperty} $\mathbf{P}_{\CODE{in}} \gets \CODE{\OOP{O}.property$(\ONTO{isIn})$}$\; 
        \OOP{B} \CODE{robotPosDescrs} $\gets$ \CODE{robotDescr.AV.build$(\mathbf{P}_{\CODE{in}})[0]$}\label{ln:ex2:buildPos}\hfill\tcp{e.g., grounded on \ONTO{Corridor1}.}
    \CODEsep%
        \tcp{Acquire knowledge about reachable locations, e.g., grounded on \ONTO{\{Room1, Room2\}}.}%
        \CODE{OWLProperty} $\mathbf{P}_{\CODE{con}} \gets \CODE{\OOP{O}.property$(\ONTO{isConnectedTo})$}$\;         
        \OOP{B}\!\!$[\,]$ \CODE{connectedPosDescrs} $\gets$ \CODE{robotPosDescr.AV.build$(\mathbf{P}_{\CODE{con}})$}\;\label{ln:ex2:connLoc}%
    \CODEsep%
        \tcp{Find the most specific representation of reachable locations.}%
        \CODE{outPairs $\gets$ $[\,]$}\;
        \ForEach{\textnormal{\CODE{\OOP{B} connection $\in$ connectedPosDescrs}}}{%
            \tcp{Acquire knowledge representing the types of reachable locations,\\ e.g., grounded on \ONTO{\{TOP, LOCATION, INDOOR, ROOM\}}.}
            \OOP{H}\!\!$[\,]$ \CODE{reachableClsDescrs} $\gets$ \CODE{connection.AC.build$()$}\label{ln:ex2:buildConn}\;%
    \CODEsep%
            \tcp{For each location descriptions, identify the most specific representation.}%
            \ForEach{\textnormal{\CODE{\OOP{H} type $\in$ reachableClsDescrs}}}{%
                \If(\label{ln:ex3:leaf}){\textnormal{\CODE{type.isLeaf$()$}}}{%
                    \CODE{outPairs.add$($\textbf{new}~Pair$($type.getGround$()$,connection.getGround$()))$}\;\label{ln:ex2:addPair}%
                    \textbf{break}%
                }%
            }%
        }%
        \Return{\textnormal{\CODE{outPairs}}}\label{ln:ex2:return}\hfill \tcp{e.g., $\{\langle\ONTO{Room1,ROOM}\rangle,\langle\ONTO{Room2,ROOM}\rangle\}.$}
\end{examplecode}

%% file: algorithm/example3D-E.tex
\begin{examplecode}[!t]%
    \caption{Two extensions of the OWLOOP compound individual descriptor implemented in Example~\ref{alg:exDesF}, which represent an individual \OOP{W} or a door \OOP{V} if the ground is \ONTO{DOOR}.}%
    \label{alg:exExt3}%
    \footnotesize%
        \SetKwBlock{ConstrV}{constructor \OOP{V}\!\!(\CODE{Ontology ontology, OWLIndividual door})}{}%
        \SetKwBlock{ConstrW}{constructor \OOP{W}\!\!(\CODE{Ontology ontology, OWLIndividual door})}{}%
        \SetKwBlock{AVBuild}{override \textnormal{\OOP{W} AV.newEntityToBuild}$($\CODE{OWLIndividual} $y_i)$}{}%
        \SetKwBlock{OpenClose}{\textnormal{\CODE{void} \CODE{updateState$($Boolean open$)$}}\label{ln:ex3D:update}}{}%
%
        \SetKwBlock{Interf}{class \textnormal{\OOP{W}} extends \textnormal{\OOP{B}}}{}%
        \SetKwBlock{Interff}{class \textnormal{\OOP{V}} extends \textnormal{\OOP{W}}}{}%
        \Interf(\label{alg:ex3D:EE}){%
            \ConstrW{%
                \SUPER{}{ontology, ground}\;%
                \CODE{OWLClass} \THIS{$\Delta_{\CODE{DOOR}}$} $\gets$ \OOP{O}.\CODE{class$($\ONTO{DOOR$)$}}
            }%
            \AVBuild(\label{ln:ex3D:doorBuild}){%
                \CODE{OWLClasses$[\,]$ types $\gets$ \OOP{O}.classify$(y_i)$}\label{ln:ex3D:OWLfactory}\;%
                \If(\label{ln:ex3:newBuild1}){\textnormal{\CODE{types.contains(\THIS{$\Delta_{\CODE{DOOR}}$})}}}{%
                    \OOP{V} \CODE{builtDescr} $\gets$ \textbf{new} \OOP{V}$(\OOP{O}, y_i)$\hfill\tcp{Construct a door descriptor \OOP{V}.}
                    \Return{\textnormal{\CODE{builtDescr}}}\label{ln:ex3:newBuild2}
                }\Else{%
                    \Return{\textnormal{\SUPER{\CODE{.build}}{$y_i$}}}\label{ln:ex3D:buildSuper}\hfill\tcp{Construct an individual descriptor \OOP{B}.}
                }
            }%
        }%
\CODEsep%
        \Interff(\label{ln:ex3D:TT}){%
            \ConstrV(\label{ln:ex3D:TTconstr}){%
                \SUPER{}{ontology, door}\;%
                \CODE{OWLClass} \THIS{$\Delta_{\CODE{OPEN}}$} $\gets$ \OOP{O}.\CODE{class$($\ONTO{OPEN$)$}}\;
                \CODE{OWLClass} \THIS{$\Delta_{\CODE{CLOSE}}$} $\gets$ \OOP{O}.\CODE{class$($\ONTO{CLOSE$)$}}
            }%
            \OpenClose(\label{ln:ex3D:openClose}){%
                \If
                {\textnormal{\CODE{open $=$ true}}}{%
                    \THIS{\CODE{AC.getEntities$()$.remove$($\THIS{$\Delta_{\CODE{CLOSE}}$}$)$}}\label{ln:ex3D:open1}\;%
                    \THIS{\CODE{AC.getEntities$()$.add$($\THIS{$\Delta_{\CODE{OPEN}}$}$)$}}\label{ln:ex3D:open2}\;%
                } \Else{
                    \THIS{\CODE{door.AC.getEntities$()$.remove$($\THIS{$\Delta_{\CODE{OPEN}}$}$)$}}\label{ln:ex3D:close1}\;%
                    \THIS{\CODE{door.AC.getEntities$()$.add$($\THIS{$\Delta_{\CODE{CLOSE}}$}$)$}}\label{ln:ex3D:close2}\;%
                }%
            }%
        }%
\end{examplecode}%

%% file: algorithm/example3.tex
\begin{examplecode}[!t]
    \caption{A script to randomly move the robot through open doors while updating the states of visited locations into the ontology shown in Figure~\ref{fig:owloop:example}.}%
    \label{alg:ex3}%
    \footnotesize%
        \SetKwBlock{Explore}{\textnormal{\CODE{void lookForDoors(Ontology \OOP{O})}}}{}\Explore{
            \tcp{Get interesting OWL entities from the ontology through IRIs.}%
            \CODE{OWLClass} $\Delta_{\CODE{DOOR}} \gets$ \OOP{O}.\CODE{class$($\ONTO{DOOR$)$}}\label{alg:ex3:const1}\;
            \CODE{OWLClass} $\Delta_{\CODE{OPEN}} \gets$ \OOP{O}.\CODE{class$($\ONTO{OPEN$)$}}\;
            \CODE{OWLClass} $\Delta_{\CODE{CLOSE}} \gets$ \OOP{O}.\CODE{class$($\ONTO{CLOSE$)$}}\;
            \CODE{OWLClass} $\rho \gets$ \OOP{O}.\CODE{individual$($\ONTO{Robot1$)$}}\;
            \CODE{OWLClass} $\mathbf{P}_{\CODE{in}} \gets$ \OOP{O}.\CODE{property$($\ONTO{in$)$}}\;
            \CODE{OWLClass} $\mathbf{P}_{\CODE{door}} \gets$ \OOP{O}.\CODE{property$($\ONTO{hasDoor$)$}}\label{alg:ex3:const2}\;
    \CODEsep%
            \tcp{Create OWL classes in the ontology to represent \ONTO{OPEN} or \ONTO{CLOSE} doors.}%
            \OOP{C} \CODE{doorStateDescr} $\gets$ \textbf{new} \OOP{C}\!\!$(\OOP{O},\Delta_{\CODE{CLOSE}})$\label{ln:ex3:ssetup}\;%
            \CODE{doorStateDescr.CS.getEntities$()$.add$(\Delta_{\CODE{DOOR}})$}\;\label{ln:ex3:close2}%
            \CODE{doorStateDescr.write$()$}\;\label{ln:ex3:close3}%
            \CODE{doorStateDescr.setGround$(\Delta_{\CODE{OPEN}})$}\;\label{ln:ex3:open1}%
            \CODE{doorStateDescr.CJ.getEntities$()$.add$(\Delta_{\CODE{CLOSE}})$}\;\label{ln:ex3:open2}%
            \CODE{doorStateDescr.write$()$}\;\label{ln:ex3:esetup}%
    \CODEsep%
            \OOP{W} \CODE{robotDescr} $\gets$ \textbf{new} \OOP{W}\!\!$(\OOP{O},\rho)$\label{ln:ex3:robot}\;%
            \While(\label{ln:ex3:loop}){\textnormal{\CODE{true}}}{%
                \tcp{Acquire robot location and the related doors.}%
                \CODE{robotDescr.read$()$}\label{ln:ex3:robotRead}\;%
                \OOP{W} \CODE{robotLocation $\gets$ robotDescr.AV.build$(\Delta_{\CODE{isIn}})[0]$}\label{ln:ex3:robotLocation}\;%
                \OOP{V}\!\!$[\,]$ \CODE{visibleDoors $\gets (\OOP{V}[\,])$ robotLocation.AV.build$(\Delta_{\CODE{hasDoor}})$}\label{ln:ex3:visibleDoor}\;%
    \CODEsep%
                \tcp{Check doors states and update the ontology accordingly.}%
                \OOP{V}\!\!$[\,]$ \CODE{openDoors} $\gets [\,]$\;
                \ForEach{\textnormal{\CODE{\OOP{V} door $\in$ visibleDoors}}}{%
                    \tcp{Manage the state of each doors.}%
                    \CODE{doorName $\gets$ door.getGround$()$.getIRI$()$}\;%
                    \CODE{Boolean openClosed $\gets$ perceiveDoorState$($doorName$)$}\label{ln:ex3:perceive}\;%
                    \CODE{door.updateState$($openClosed$)$}\label{ln:ex3:updateState}\;
                    \CODE{door.write$()$}\label{ln:ex3:updateStateWrite}\;
                    \If{\textnormal{\CODE{openClosed $=$ true}}}{%
                        \CODE{openDoors.add$($door$)$}\label{ln:ex3:openDoors}\;%
                    }%
                }%
    \CODEsep%
                \tcp{Move the robot.}%
                \CODE{Integer rdmIdx $\gets$ randomValue$($openDoors.size$())$}\label{ln:ex3:smove}\;
                \CODE{\OOP{V} nextDoorDescr $\gets$ openDoors$[$rdmIdx$]$}\;
                \CODE{moveRobotThrough$($\OOP{O}, nextDoorDescr.getGround$()$.getIRI$())$}\label{ln:ex3:emove}\;  
            }%
        }%
    \CODEsep%
        \SetKwBlock{Perceives}{\textnormal{\CODE{Boolean perceiveDoorState$($IRI doorName$)$}}\label{ln:ex3:perceiveDef}}{}%
        \Perceives{$\ldots$\hfill\tcp{Returns true if the given door is open, false if it is closed.}}
        \SetKwBlock{Moves}{\textnormal{\CODE{void moveRobotThrough$($Ontology \OOP{O}, IRI doorName$)$}}\label{ln:ex3:move}}{}%
        \Moves{$\ldots$\hfill\tcp{~~~Blocking call that moves the robot through a target door and updates\,}\hfill\tcp{its position in the ontology, which is retrieved at Line~\ref{ln:ex3:robotRead} and \ref{ln:ex3:robotLocation}.}}%
        \SetKwBlock{Random}{\textnormal{\CODE{Integer randomValue$($Integer maxValue$)$}}}{}%
        \Random{$\ldots$\hfill\tcp{Returns a random value spanning in $[0,\CODE{maxValue})$.}}%
\end{examplecode}